\documentclass[conference]{IEEEtran}
\IEEEoverridecommandlockouts

\usepackage{cite}
\usepackage{amsmath,amssymb,amsfonts}
\usepackage{algorithmic}
\usepackage{graphicx}
\usepackage{textcomp}
\usepackage{xcolor}
\usepackage{float}
\usepackage{booktabs}
\usepackage{multirow}
\usepackage[colorlinks=true, linkcolor=blue, citecolor=blue, urlcolor=blue]{hyperref}
\usepackage{enumitem}
\usepackage{colortbl}
\usepackage{subcaption}

\usepackage{graphicx}
\usepackage{array}
\usepackage{makecell} 
\usepackage{cellspace}

\usepackage{array}  
\usepackage{graphicx} 

\definecolor{lightgreen}{rgb}{0.9,1,0.9}

\makeatletter
\let\IEEE@orig@ps@title\ps@IEEEtitlepagestyle
\def\ps@IEEEtitlepagestyle{%
  \IEEE@orig@ps@title 
  \def\@oddhead{\parbox{\textwidth}{\centering\normalsize
    2025 7\textsuperscript{th} International Conference on Pattern Recognition and Image Analysis (IPRIA), 24--25 September, 2025,\\
    Islamic Azad University--Lahijan Branch, Lahijan, Iran}}%
  \def\@evenhead{\@oddhead}%
}
\makeatother

\def\BibTeX{{\rm B\kern-.05em{\sc i\kern-.025em b}\kern-.08em
    T\kern-.1667em\lower.7ex\hbox{E}\kern-.125emX}}

\IEEEpubid{\makebox[\columnwidth]{979-8-3315-5846-8/25/\$31.00~\copyright~2025 IEEE\hfill}
\hspace{\columnsep}\makebox[\columnwidth]{}}

\usepackage[T1]{fontenc}

\title{Enhancing 3D Point Cloud Classification with ModelNet-R and Point-SkipNet }

\makeatletter
\def\@IEEEauthorblockNstyle{\normalfont\fontsize{9pt}{11pt}\selectfont}
\def\@IEEEauthorblockAstyle{\normalfont\fontsize{9pt}{11pt}\selectfont}
\makeatother

\author{
\IEEEauthorblockN{Mohammad Saeid}
\IEEEauthorblockA{
\textit{Sirjan University of Technology}\\
Sirjan, Iran \\
m.saeid@stu.sirjantech.ac.ir}
\and
\IEEEauthorblockN{Amir Salarpour}
\IEEEauthorblockA{
\textit{Sirjan University of Technology}\\
Sirjan, Iran \\
salarpour@sirjantech.ac.ir}
\and
\IEEEauthorblockN{Pedram MohajerAnsari}
\IEEEauthorblockA{
\textit{Clemson University}\\
Clemson, SC, USA \\
pmohaje@clemson.edu}
}

\begin{document}

\maketitle

\begin{abstract}
The classification of 3D point clouds is crucial for applications such as autonomous driving, robotics, and augmented reality. However, the commonly used ModelNet40 dataset suffers from limitations such as inconsistent labeling, 2D data, size mismatches, and inadequate class differentiation, which hinder model performance. This paper introduces ModelNet-R, a meticulously refined version of ModelNet40 designed to address these issues and serve as a more reliable benchmark. Additionally, this paper proposes Point-SkipNet, a lightweight graph-based neural network that leverages efficient sampling, neighborhood grouping, and skip connections to achieve high classification accuracy with reduced computational overhead. Extensive experiments demonstrate that models trained in ModelNet-R exhibit significant performance improvements. Notably, Point-SkipNet achieves state-of-the-art accuracy on ModelNet-R with a substantially lower parameter count compared to contemporary models. This research highlights the crucial role of dataset quality in optimizing model efficiency for 3D point cloud classification. For more details, see the code at: \url{https://github.com/m-saeid/ModeNetR_PointSkipNet}.

\end{abstract}

\IEEEpubidadjcol


\section{Introduction}
\label{sec:intro}
Rapid advancements in 3D scanning and sensor technologies have made the acquisition of detailed point cloud data more accessible, leading to significant progress in various applications such as autonomous driving \cite{mohajeransari2024discovering, fernandez2025avoiding}, robotics \cite{ wang2020grasping, salarpour2014long}, and augmented reality \cite{mahmood2020bim, jin2020mobile, lim2022point}. Despite their growing availability, point clouds are inherently high-dimensional and unstructured, posing unique challenges for robust data processing \cite{li2024graph }. These challenges highlight the importance of not only developing powerful classification models, but also ensuring that the underlying datasets are accurate and consistent.

\begin{figure}[t] 
    \centering
    \begin{subfigure}[b]{0.32\columnwidth}
        \centering
        \includegraphics[width=\linewidth]{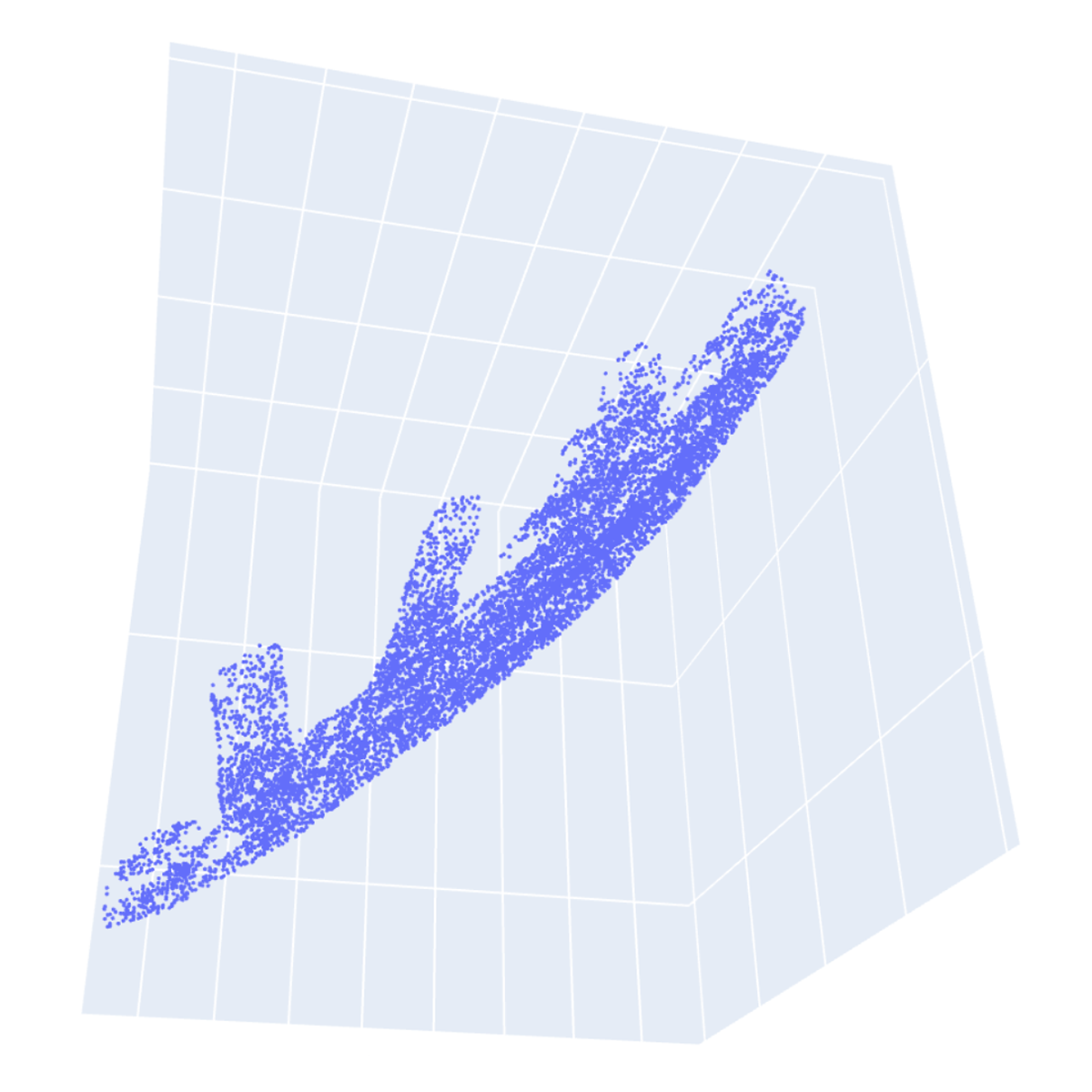}
        \caption{2D data}
    \end{subfigure}
    \hfill
    \begin{subfigure}[b]{0.32\columnwidth}
        \centering
        \includegraphics[width=\linewidth]{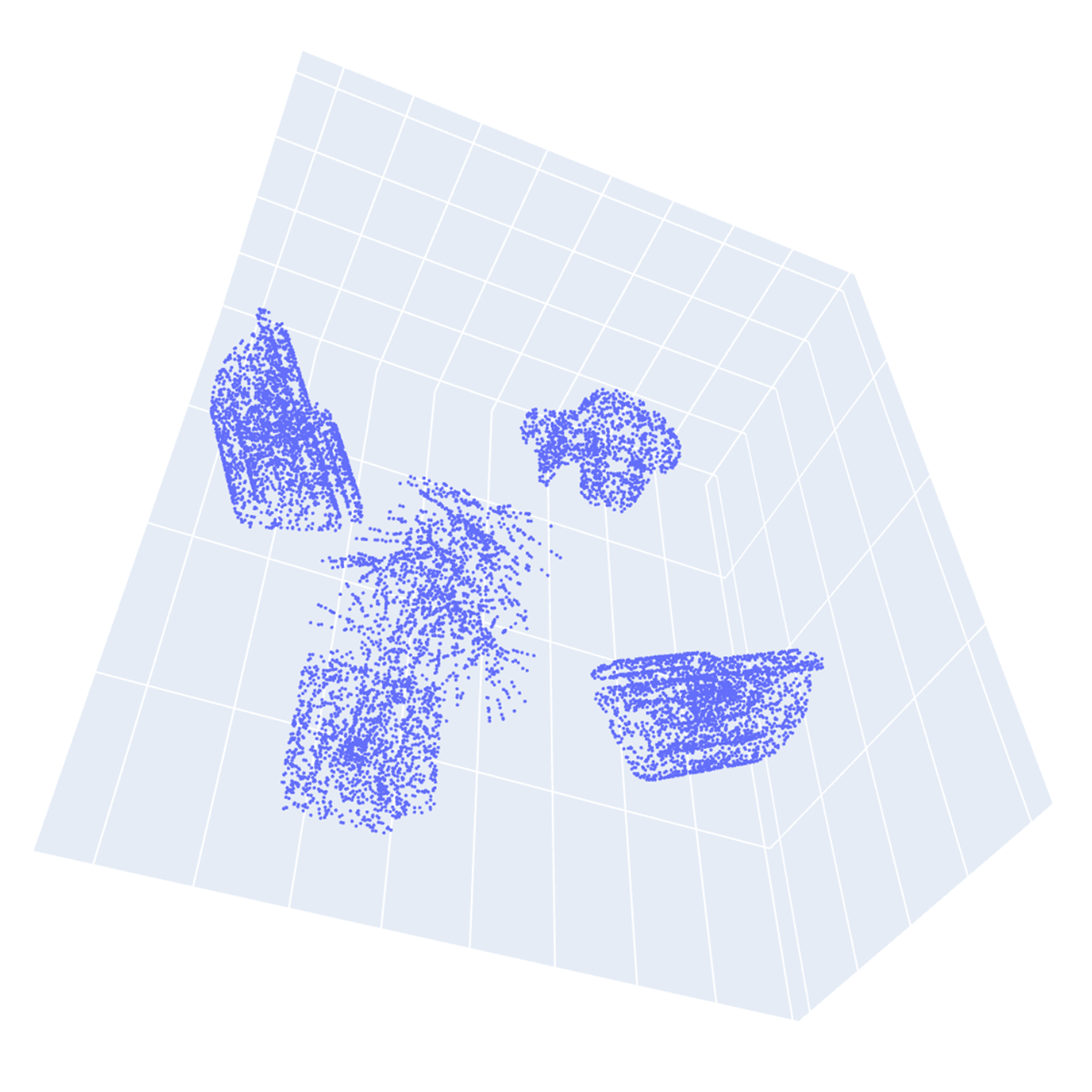} 
        \caption{Multi object}
    \end{subfigure}
    \hfill
    \begin{subfigure}[b]{0.32\columnwidth}
        \centering
        \includegraphics[width=\linewidth]{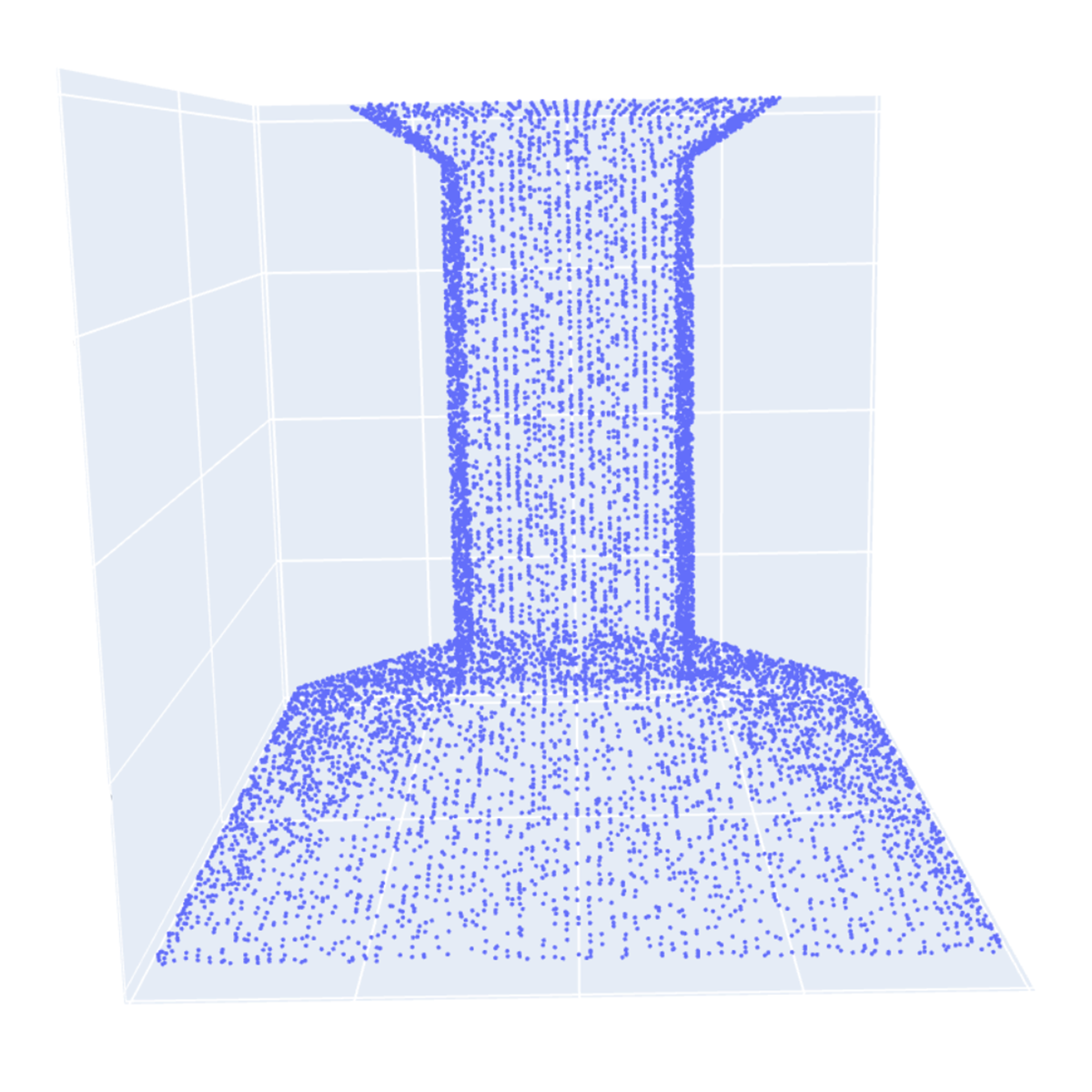} 
        \caption{Wrong label}
    \end{subfigure}
    \caption{Some problems of ModelNet dataset.}
    \label{fig:sample_problem}
    \vspace{-5mm}
\end{figure}

Although the ModelNet \cite{Wu_2015_CVPR} dataset is widely used for benchmarking 3D point cloud classification, it suffers from various issues such as inconsistent labeling, 2D or low-quality data, and lacking class differentiation. These issues can undermine the reliability of training data and degrade model performance. \autoref{fig:sample_problem} shows some of the problems, including instances of misclassified objects and datasets that cannot be reliably classified.

\begin{figure*}[t]
    \centering
    \includegraphics[width=1.9\columnwidth]{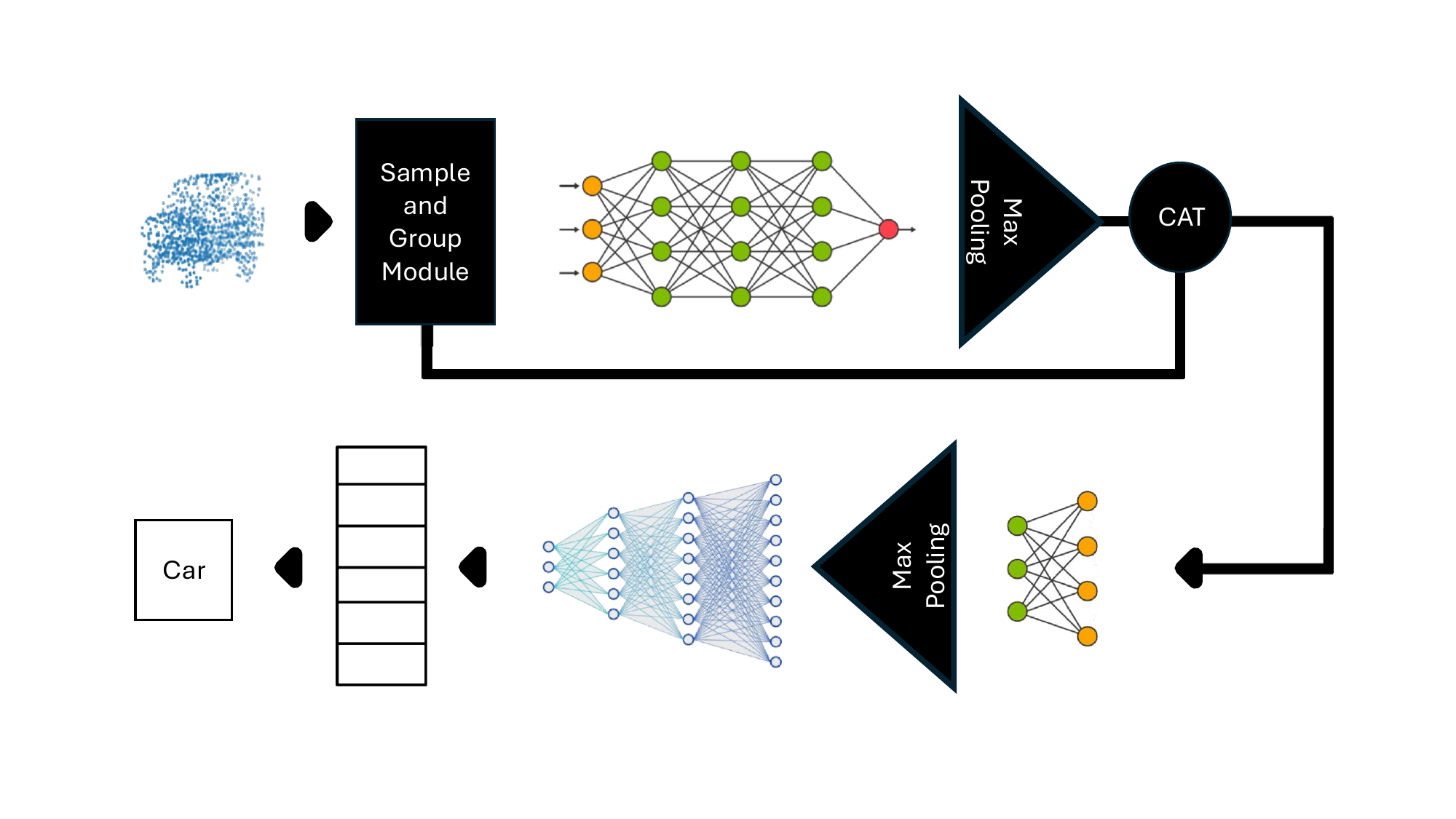}
    \vspace{-10mm}
    \caption{General architecture of Point-SkipNet}
    \label{fig:Abstract}
    \vspace{-5mm}
\end{figure*}

Existing point-based models like PointMLP \cite{ma2022rethinking} and APES \cite{wu2023attention} have set impressive benchmarks for point cloud classification. However, they are often computationally intensive, making them less suitable for resource-constrained environments \cite{aldeeniv}. Hence, there is a growing need for lighter, more efficient architectures that can still achieve competitive accuracy. In parallel, relatively few works have addressed the fundamental quality issues in popular datasets, instead focusing on either expanding them or creating new benchmarks.

Motivated by these gaps, this study aims to improve both the data and the model. First, we refine the ModelNet dataset to create \textbf{ModelNet-R}, a cleaner and more consistent resource for point cloud research. This refined version removes mislabeled entries, eliminates 2D data, and strengthens class distinctions. Second, we propose a lightweight graph-based neural network architecture called \textbf{Point-SkipNet}, tailored for efficient yet accurate classification of 3D point clouds \autoref{fig:Abstract} gives an overview of the \textbf{Point-SkipNet} architecture, showing how it leverages graph-based operations to reduce computational overhead while preserving key geometric features.

Extensive experiments show that ModelNet-R significantly improves the classification accuracy across various models, including our proposed \textbf{Point-SkipNet}. In particular, \textbf{Point-SkipNet} achieves state-of-the-art performance on ModelNet-R with much lower computational cost than many of its competitors, highlighting the importance of both dataset quality and efficient network design. In summary, our key contributions include:

\begin{itemize}
    \item \textbf{ModelNet-R Dataset}: Refining the ModelNet by resolving labeling inconsistencies, removing low-quality data, and improving class definitions.
    \item \textbf{Point-SkipNet Model}: Introducing Point-SkipNet, a lightweight graph-based architecture designed for efficient and accurate point cloud classification, particularly in resource-limited environments.
    \item \textbf{Performance Evaluation}: Evaluating the impact of dataset quality and model efficiency on 3D classification accuracy, demonstrating their combined potential for improved performance.
\end{itemize}


\section{Related works}
\label{sec:related_work}

\subsection{3D Point Cloud Datasets}
Over the years, numerous 3D datasets have been introduced to drive research in areas such as autonomous driving \cite{li2020deep, cui2021deep, abbasi2022lidar}, robotics \cite{duan2021robotics, wang2020grasping, wang2021trajectory}, and augmented reality \cite{mahmood2020bim, jin2020mobile, lim2022point}. These datasets encompass diverse real-world conditions, from large-scale outdoor driving environments to cluttered indoor spaces for service robots, driving significant advancements in 3D perception. However, many suffer from limited diversity, poor real-world representation, or inconsistent data quality and annotations.

Researchers have responded to these shortcomings in a variety of ways. \textbf{ShapeNet} \cite{chang2015shapenet} expanded the idea of CAD-based datasets by collecting over 3~million 3D models spanning more than 3{,}000 categories, though many of these categories have limited samples. \textbf{ScanObjectNN} \cite{uy2019revisiting} addressed the gap between synthetic CAD models and real-world scans by offering 2{,}902 unique objects from 15 categories, captured under noisy and occluded conditions. \textbf{S3DIS} \cite{armeni20163d} focused on large-scale indoor scenes, offering dense 3D scans of office environments, while \textbf{SUN RGB-D} \cite{song2015sun} provided RGB-D imagery for over 10{,}000 indoor scenes with instance segmentation. In the realm of autonomous driving, \textbf{nuScenes} \cite{caesar2020nuscenes} and \textbf{KITTI} \cite{geiger2013vision} are prominent benchmarks for capturing outdoor driving scenarios with detailed annotations for objects and trajectories.

Despite the introduction of diverse datasets, ModelNet remains a key benchmark for point cloud classification, valued for its accessibility and standardized training protocols. Thus, improving the consistency and overall quality of ModelNet can have a broad impact on the field, benefiting researchers who rely on it for model validation and comparison.

\subsection{3D Point Cloud Classification}

Early approaches to point cloud classification often transformed 3D data into alternative representations. \textbf{Voxel-based methods} \cite{zhou2018voxelnet} discretized space into 3D grids, preserving structure but suffering from high memory costs and loss of fine-grained details. \textbf{Projection-based methods} \cite{ahn2022projection, jhaldiyal2023semantic, gopi2002fast} mapped point clouds onto 2D planes for convolutional neural network (CNN) processing \cite{lecun1989backpropagation}, although these projections could omit critical 3D information.

To overcome these issues, \textbf{point-based} architectures have emerged as the dominant paradigm. \textbf{PointNet} \cite{qi2017pointnet} pioneered direct point cloud processing employing symmetry functions and multilayer perceptrons (MLPs), while \textbf{PointNet++} \cite{qi2017pointnet++} introduced a hierarchical feature extraction strategy. Building on these ideas, \textbf{DGCNN} \cite{wang2019dynamic} applied dynamic graph constructions to capture local neighborhood information, and \textbf{KPConv} \cite{thomas2019kpconv} used kernel point convolutions for improved local feature learning. More recently, \textbf{transformer-based} models \cite{zhao2021point, wu2022point, wu2024point, guo2021pct} have leveraged attention mechanisms to capture global dependencies among points, achieving strong performance across various datasets, albeit at a high computational cost.

In parallel, non-parametric methods have been explored for robust classification by adaptively modeling local neighborhoods under varying data distributions \cite{zhang2023parameter, salarpour2024pointgn, salarpour2025pointln}. Furthermore, plane-based approaches such as Point-PlaneNet \cite{peyghambarzadeh2020point} leverage plane kernels to better capture geometric context, improving the classification accuracy in complex point cloud shapes.

\subsection{Research Gaps and Proposed Solutions}

Despite substantial advancements in both datasets and classification approaches, two overarching issues remain.

\begin{enumerate}[label=(\roman*)] 
    \item Dataset Quality and Consistency: While newer datasets have improved real-world complexity representation, ModelNet continues to be a fundamental benchmark in 3D research. It continues to feature mislabeled, incomplete, or 2D data that can degrade model performance and benchmark reliability.
    \item Computational Efficiency: Many top-performing models (e.g., transformer-based architectures) demand significant computational resources, limiting their usability in real-time or low-power settings (such as embedded systems or mobile devices).
\end{enumerate} 

In this work, we address these challenges by introducing ModelNet-R, a refined variant of ModelNet that removes mislabeled/2D entries and clarifies ambiguous class distinctions.
Additionally, we propose Point-SkipNet, a lightweight graph-based classification network that balances accuracy with computational efficiency, making it suitable for real-time and resource-constrained applications.

\section{Methodology}
\label{sec:methodology}

\subsection{From ModelNet to ModelNet-R: Enhancing dataset reliability and model performance}
\label{subsec:modelnetr_overview}

\begin{figure}[t]
    \centering
    \subfloat{\includegraphics[width=0.12\textwidth]{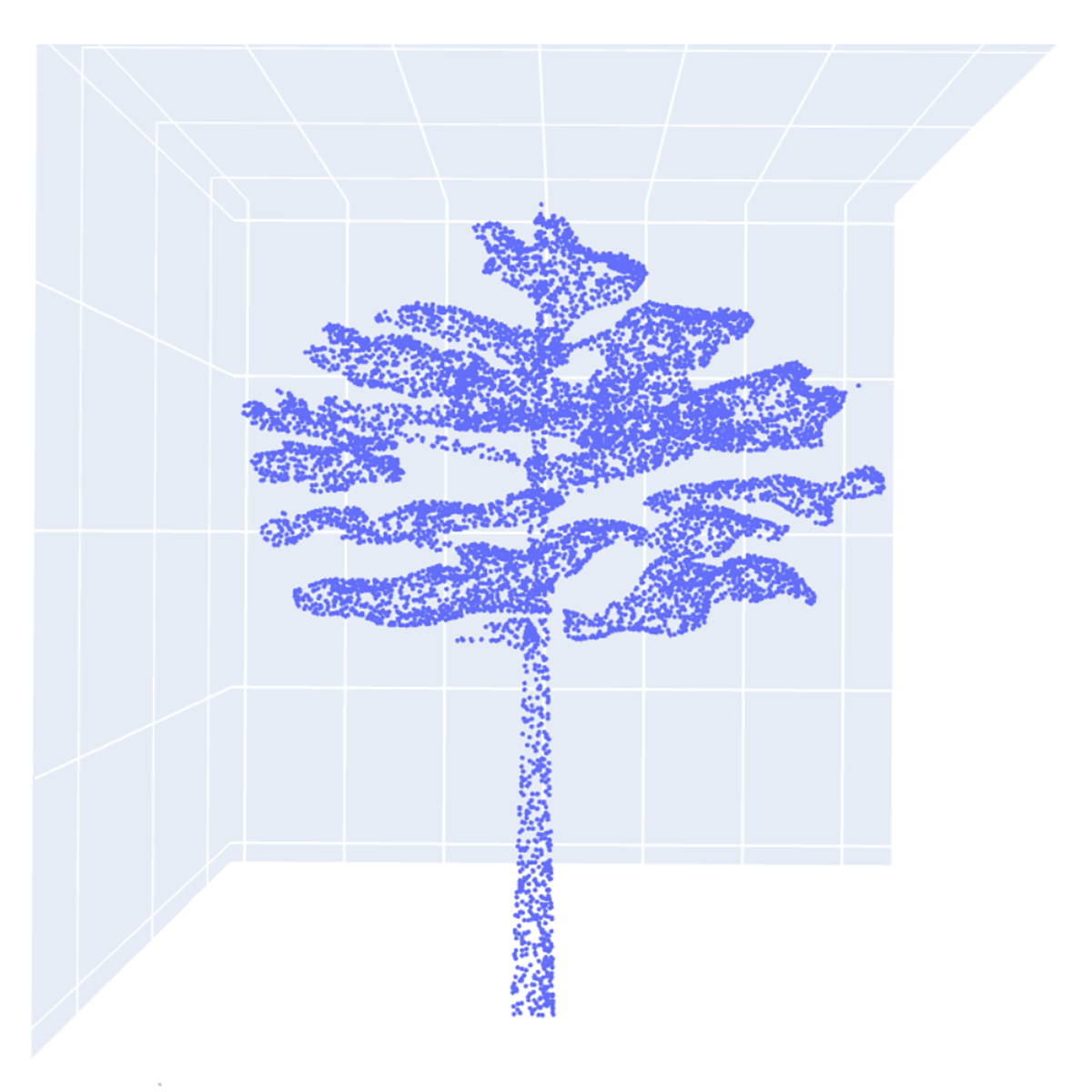}}
    \subfloat{\includegraphics[width=0.12\textwidth]{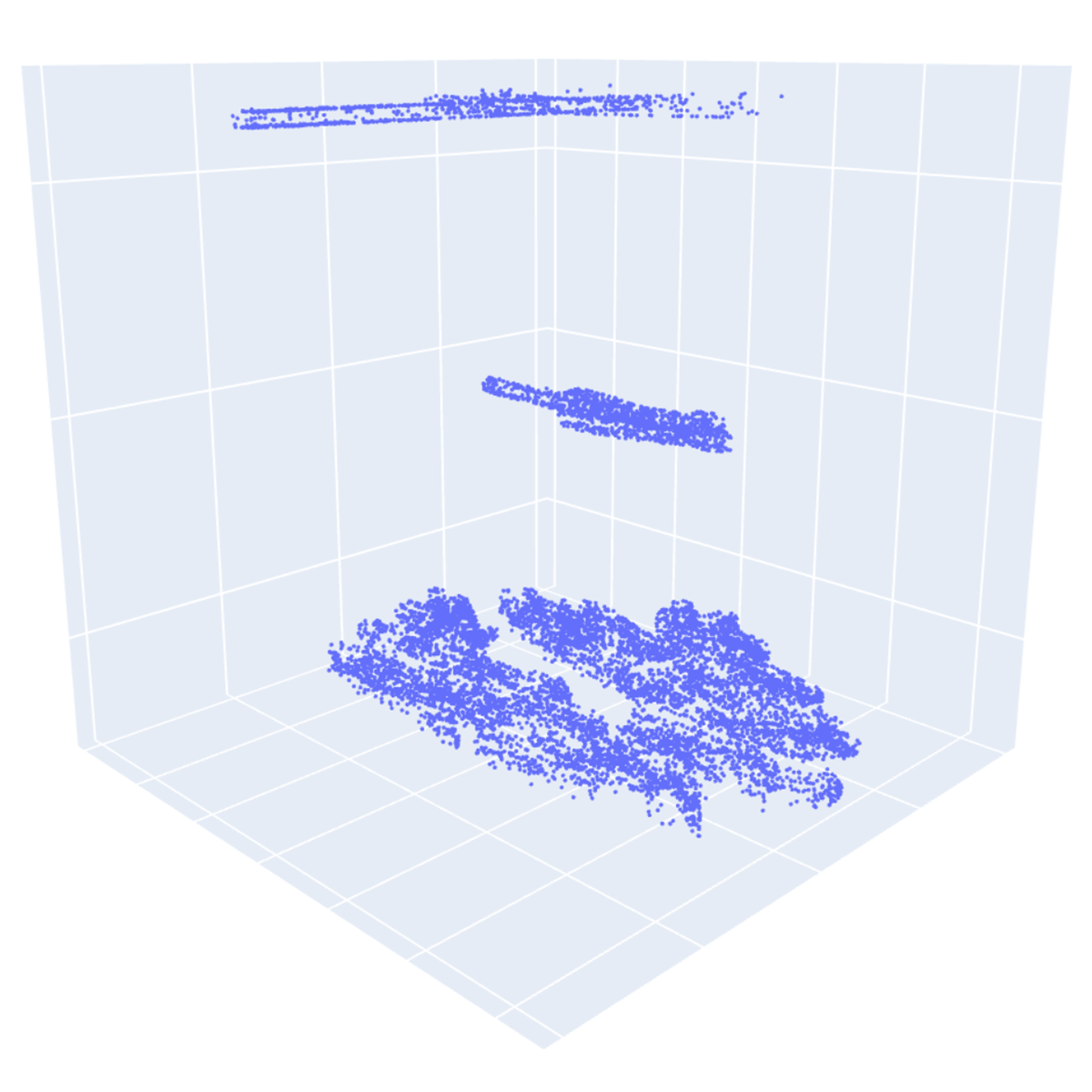}}
    \subfloat{\includegraphics[width=0.12\textwidth]{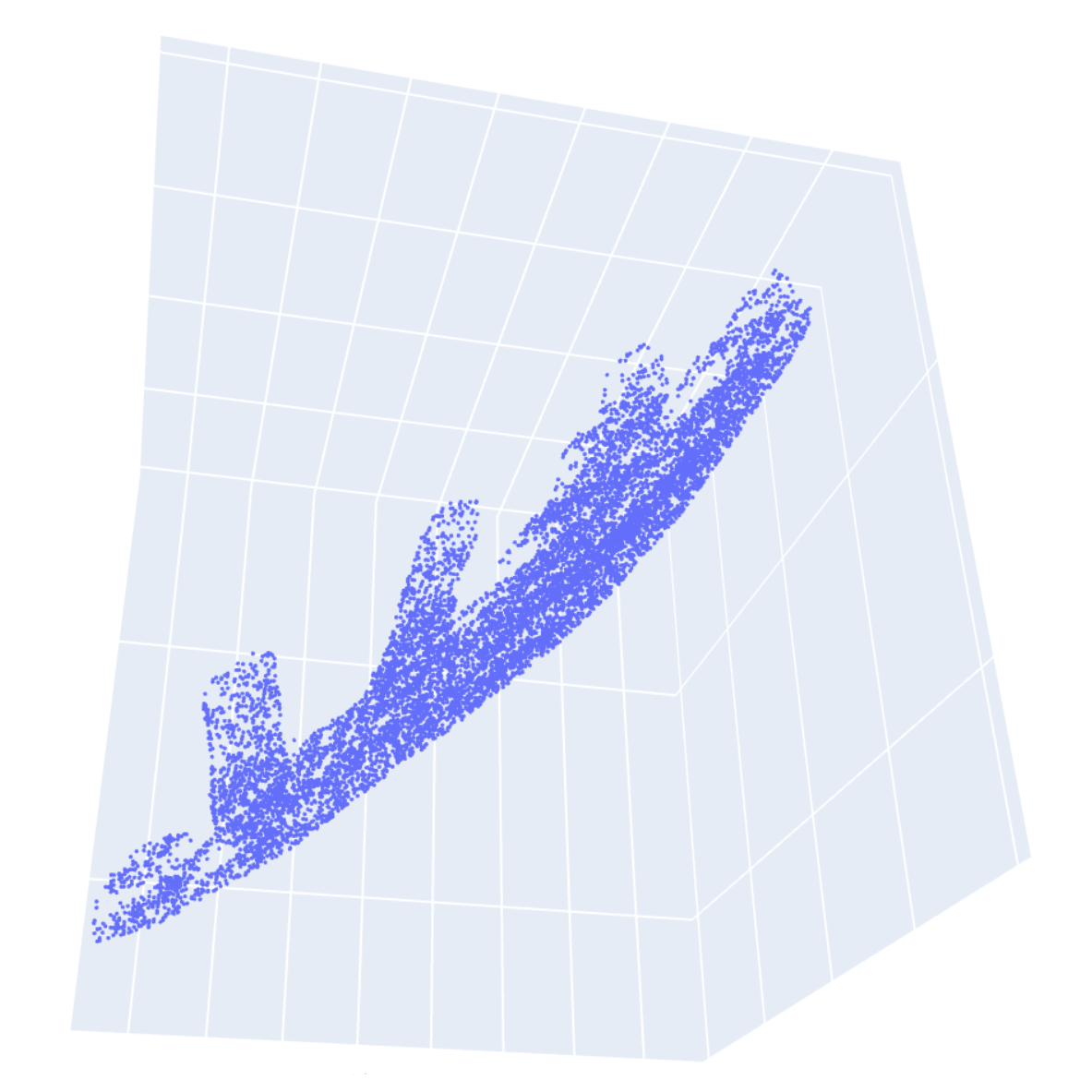}}
    \subfloat{\includegraphics[width=0.12\textwidth]{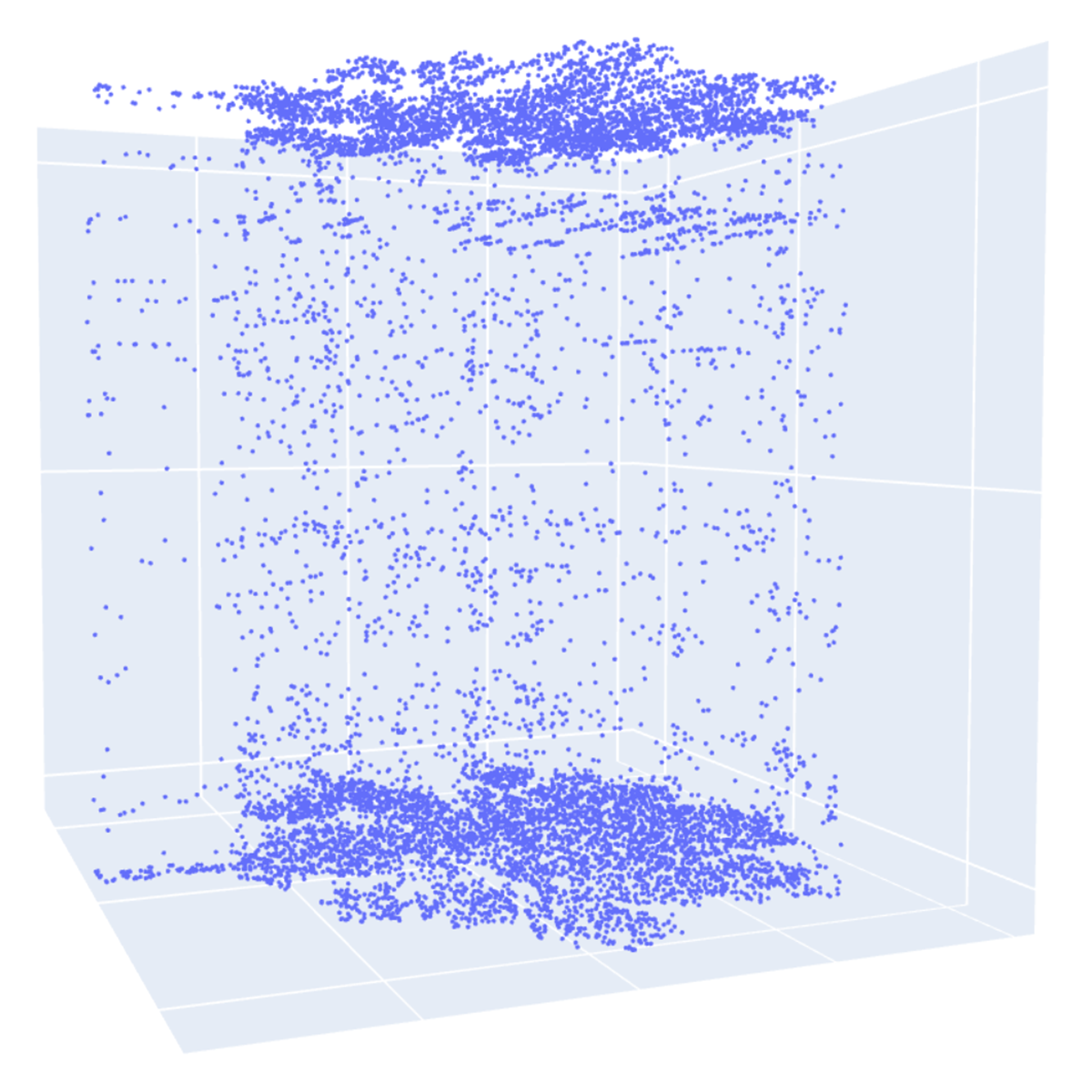}}

    \vspace{2mm}

    \subfloat{\includegraphics[width=0.12\textwidth]{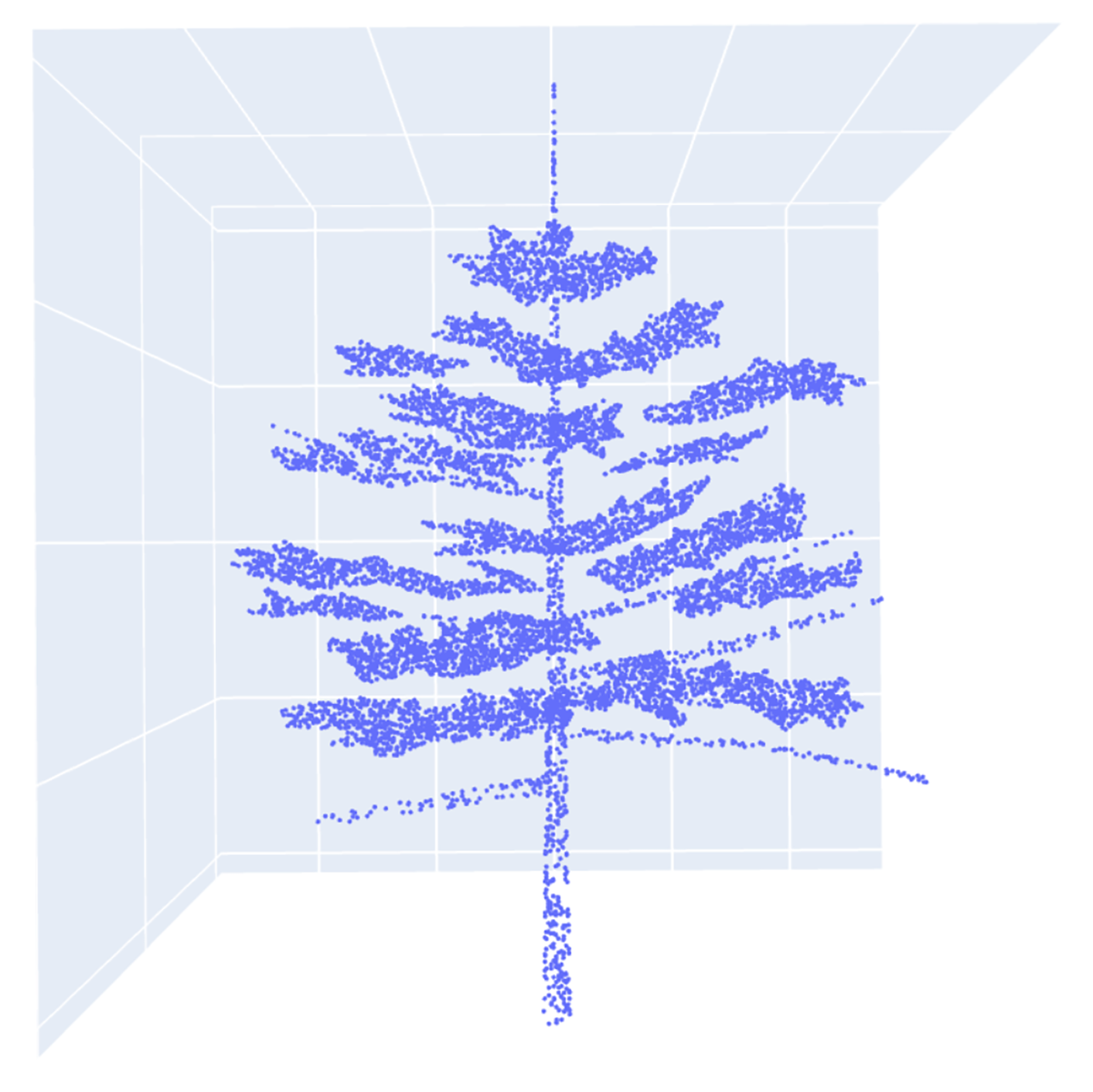}}
    \subfloat{\includegraphics[width=0.12\textwidth]{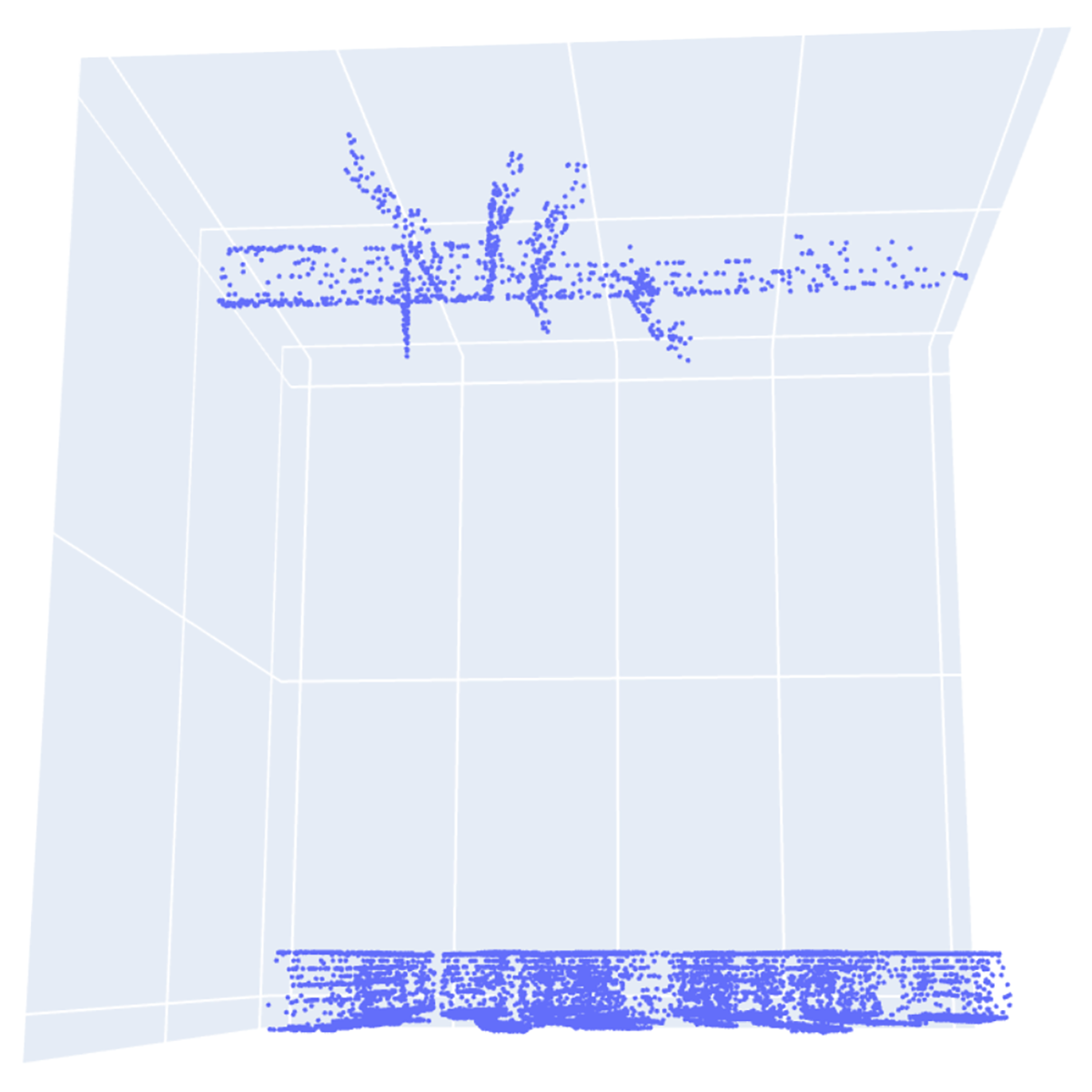}}
    \subfloat{\includegraphics[width=0.12\textwidth]{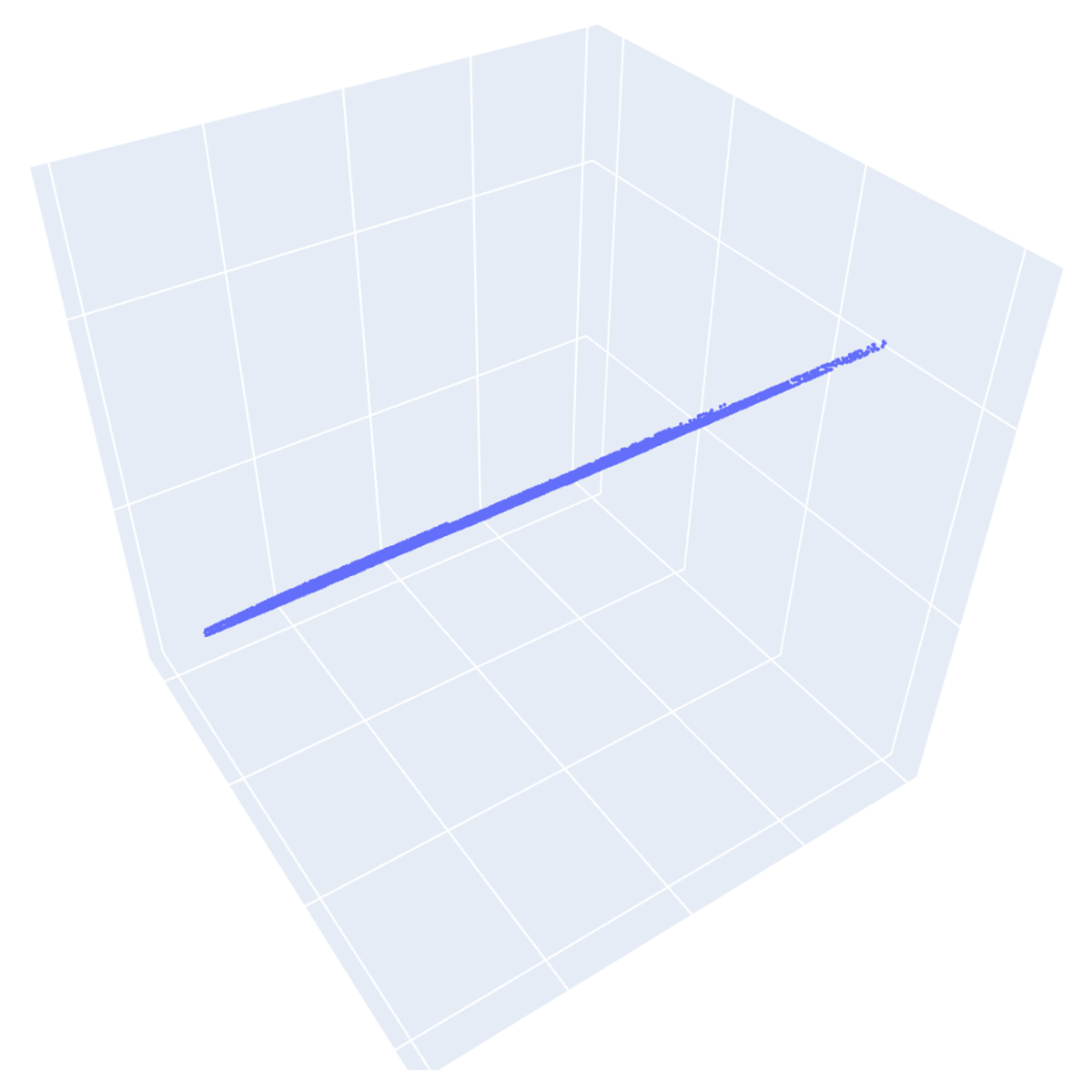}}
    \subfloat{\includegraphics[width=0.12\textwidth]{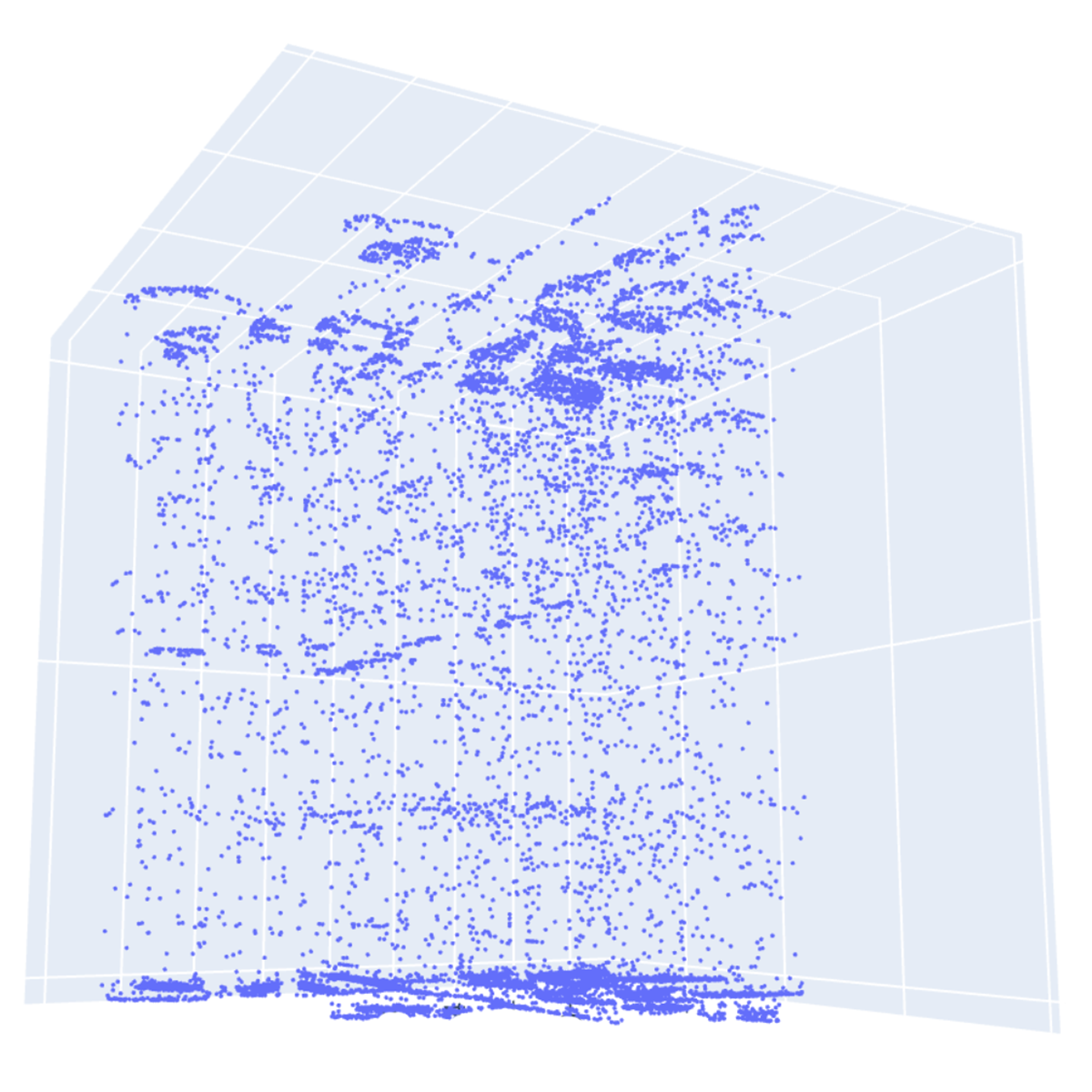}}
    \caption{Flattened 2D structures in ModelNet lacking volumetric depth.}

    \label{fig:2d}
    \vspace{-5mm}
\end{figure}

This section details the transformation of the original ModelNet dataset into \textbf{ModelNet-R}, a refined version that addresses issues such as inconsistent labeling, 2D data, size mismatches, and overlapping class definitions. \autoref{table:statis_ModelNet} summarizes the initial statistical characteristics of the ModelNet dataset. By systematically correcting labels, removing problematic samples, and adjusting class definitions, \textbf{ModelNet-R} becomes a more reliable benchmark for \textit{3D point cloud classification}, ultimately improving both training efficacy and model performance. These issues include inconsistent labeling, the presence of 2D data, data size mismatches, and overlapping class definitions, as detailed below:

\begin{table}[h]
    \centering
    \footnotesize
    \renewcommand{\arraystretch}{1.1}

    \caption{Key statistics of the ModelNet dataset.}
    \begin{tabular}{|c|c|}
        \hline
        Total Number of Classes & 40 \\
        \hline
        Total Number of Instances & 12,311 \\
        \hline
        Average Instances per Class & 307.775 \\
        \hline
        Class with Maximum Instances & 989 \\
        \hline
        Class with Minimum Instances & 84 \\
        \hline
    \end{tabular}
    \label{table:statis_ModelNet}
    \vspace{-2mm}

\end{table}

\subsubsection{Inconsistent Labeling}
\label{subsubsec:inconsistent_labeling}

Mislabeled samples were identified through visual inspection and expert cross-referencing. When the correct label was discernible, we reassigned it. If a sample was too ambiguous (e.g., containing multiple objects) or unrecognizable, it was removed from the dataset. \autoref{fig:wrong_labels} shows examples of these label corrections.

\subsubsection{Two-Dimensional Data}
\label{subsubsec:two_dim_data}

Many samples in ModelNet were low-quality and almost flat, existing on a single plane rather than being true 3D structures. Due to the lack of volumetric depth, these samples were removed. \autoref{fig:2d} illustrates examples of these 2D samples.

\begin{table}[t]

\centering
\caption{\small Examples of misclassified samples in ModelNet.}
\renewcommand{\arraystretch}{1.5} 
\setlength{\tabcolsep}{4pt}      
\resizebox{\columnwidth}{!}{%
\begin{tabular}{|>{\centering\arraybackslash}m{0.09\textwidth}
                    |>{\centering\arraybackslash}m{0.09\textwidth}
                    |>{\centering\arraybackslash}m{2.5cm}
                    |>{\centering\arraybackslash}m{2.5cm}|}
    \hline
    \multicolumn{2}{|c|}{\textbf{Instance}} & \textbf{Modified Label} & \textbf{Label in the Dataset} \\
    \hline
    \includegraphics[width=0.08\textwidth]{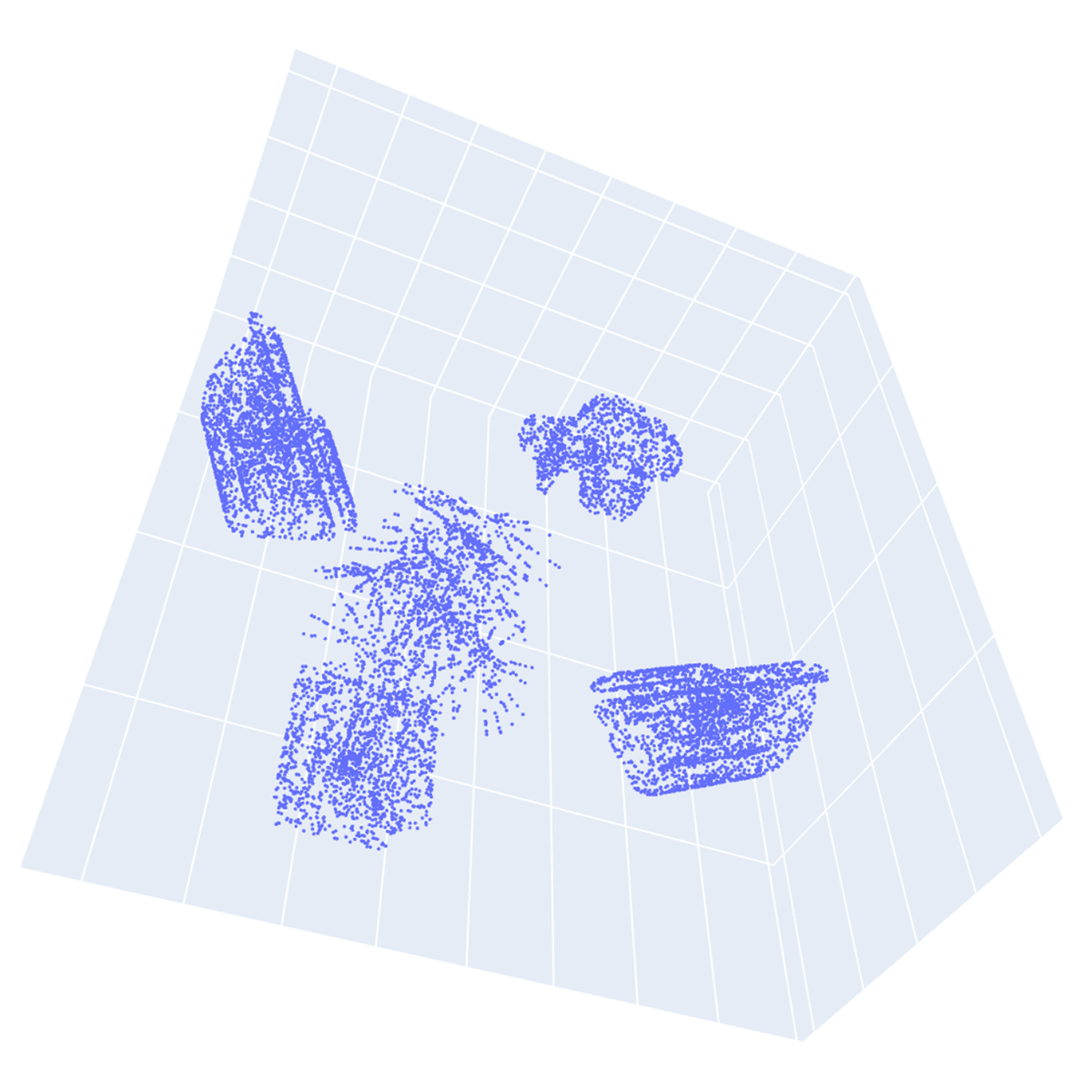} &
    \includegraphics[width=0.08\textwidth]{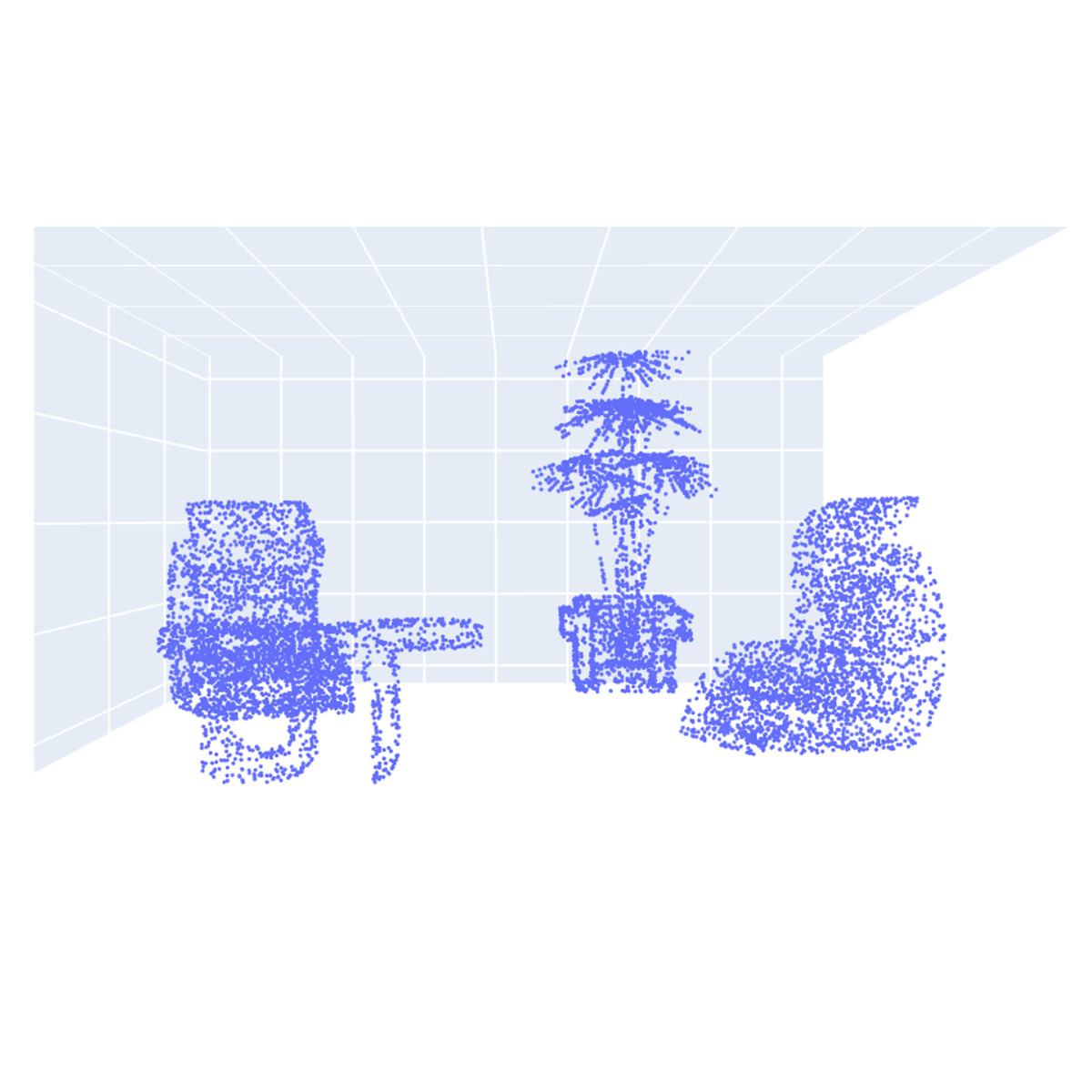} &
    Remove & Plant \\
    \hline
    \includegraphics[width=0.08\textwidth]{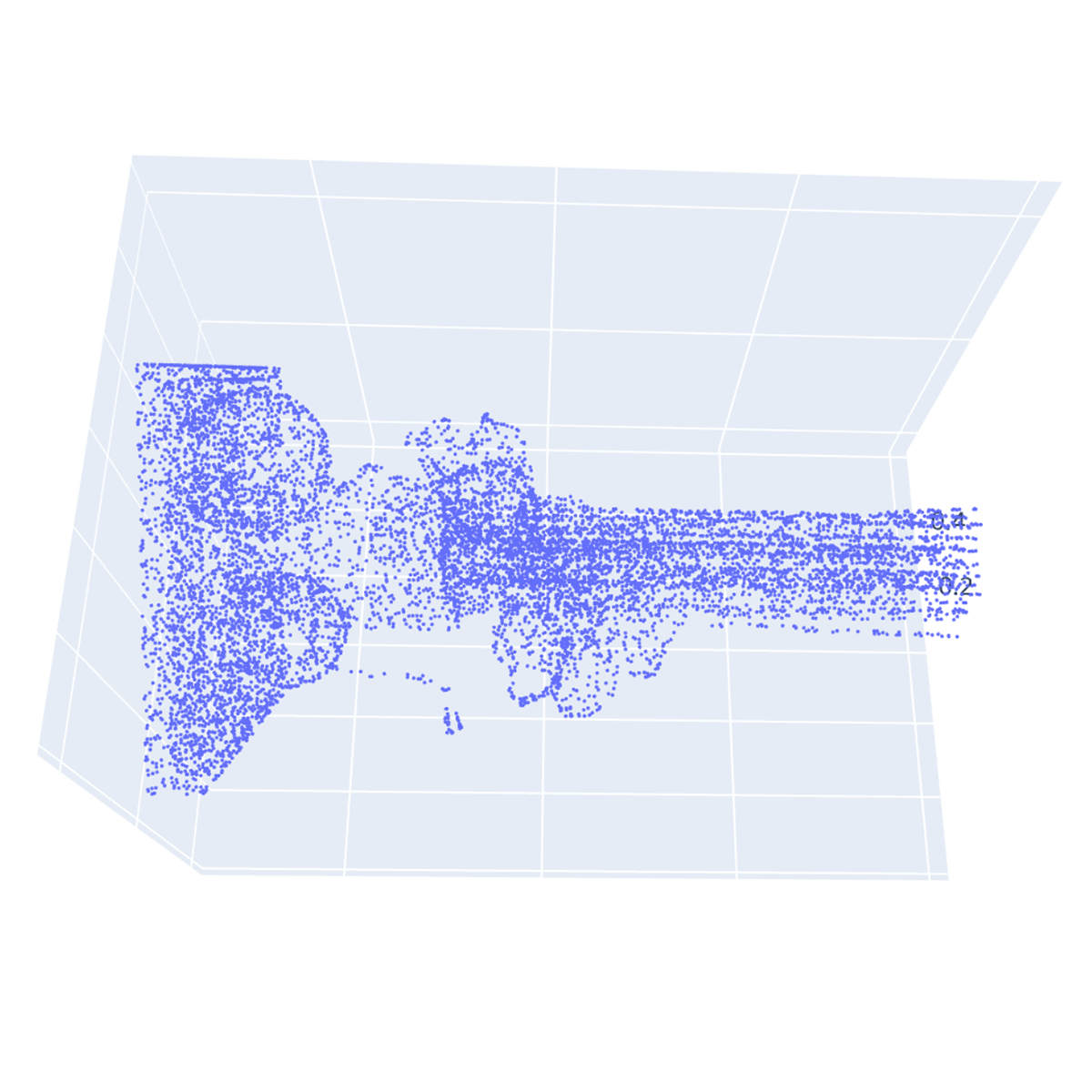} &
    \includegraphics[width=0.08\textwidth]{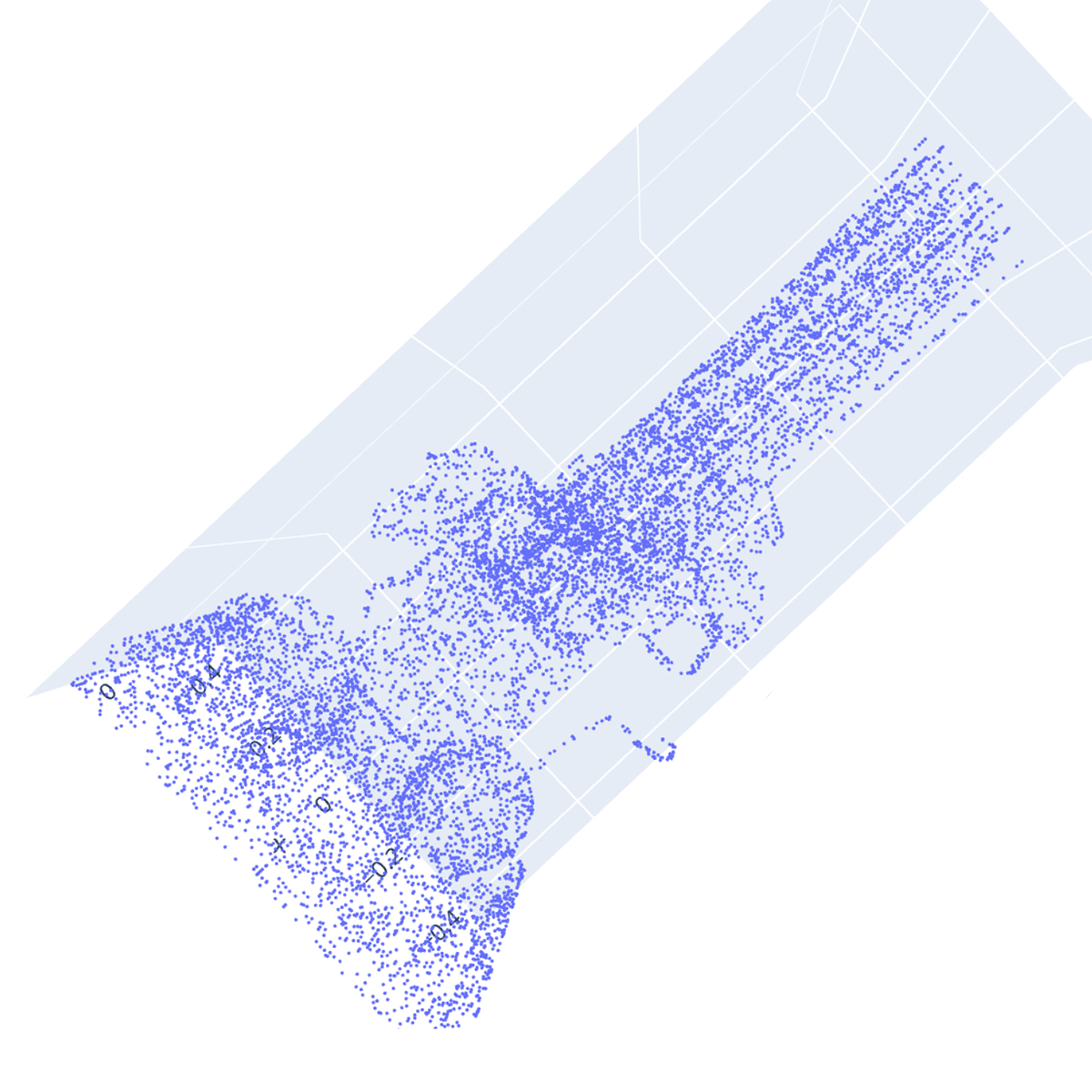} &
    Remove & Plant \\
    \hline
    \includegraphics[width=0.08\textwidth]{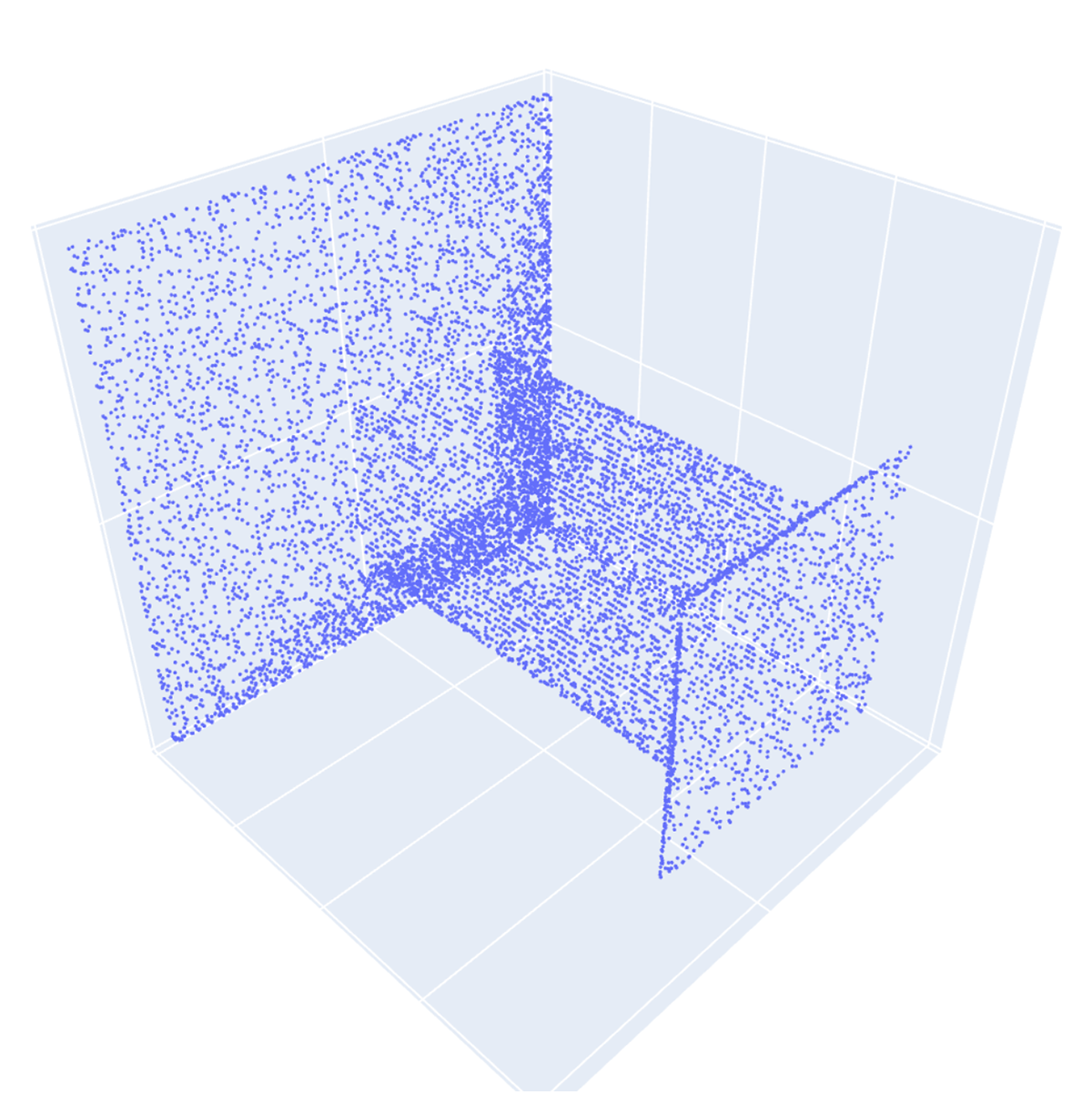} &
    \includegraphics[width=0.08\textwidth]{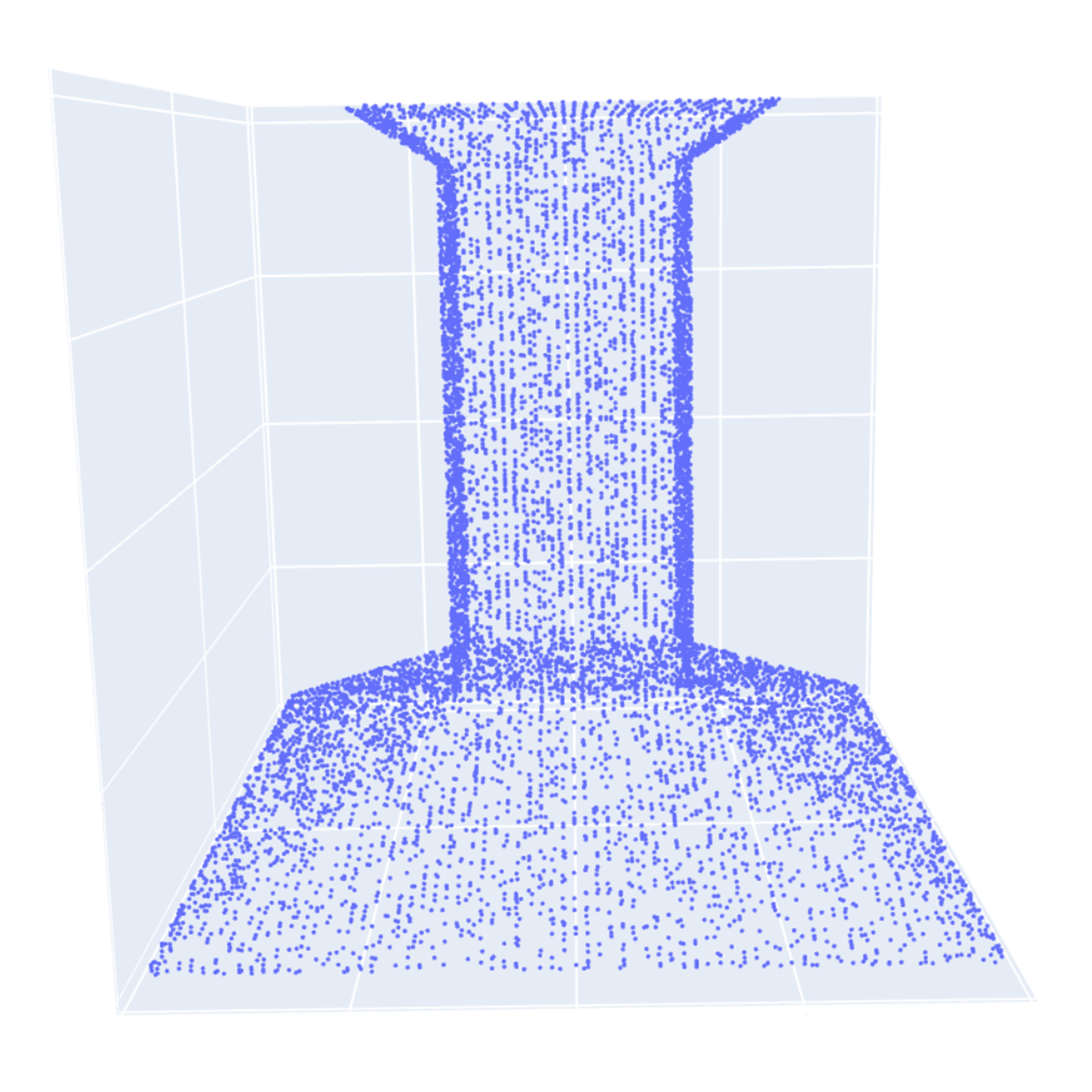} &
    Range hood & Vase \\
    \hline
\end{tabular}
}
\label{fig:wrong_labels}
\vspace{-5mm}
\end{table}

\begin{figure}[b]
    \centering
    \includegraphics[width=1\columnwidth]{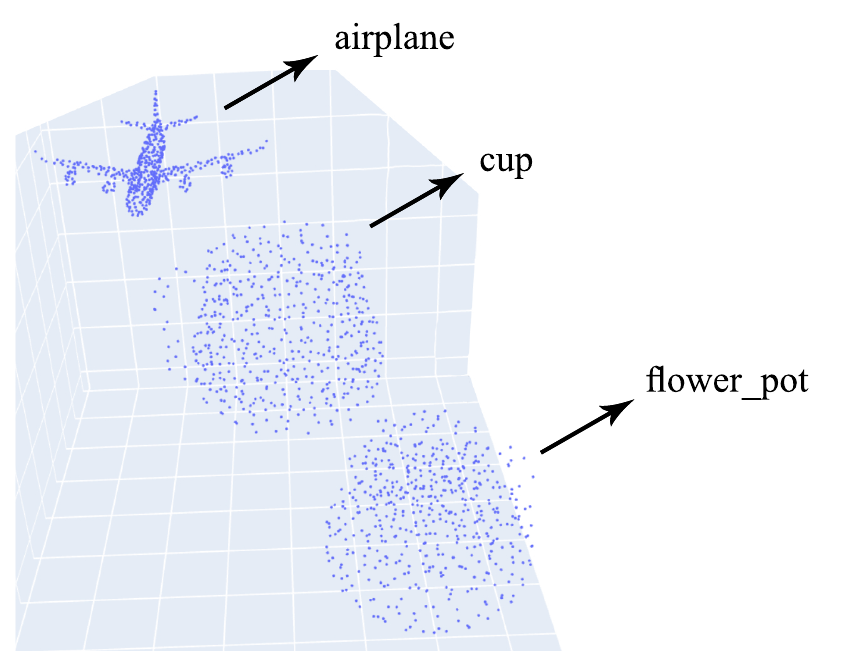}
    \caption{Size inconsistencies: normalized objects appear similar despite real-world scale differences.}
    \vspace{-3mm}
    \label{fig:mismatch}
\end{figure}

When normalized, objects with drastically different real-world sizes (e.g., airplanes and cups) can appear nearly on the same scale, causing classification confusion. To mitigate this issue, we refined class definitions, ensuring that models focus on meaningful geometric features instead of relying on size alone. \autoref{fig:mismatch} highlights these inconsistencies, highlighting the need for class adjustments that are sensitive to size.

\subsubsection{Lack of Differentiation}
\label{subsubsec:lack_of_diff}

Some classes shared highly similar geometries, making them difficult for models to distinguish (e.g., 'flower\_pot' vs. 'vase'). We adjusted ambiguous classes to ensure clearer boundaries:

\begin{itemize}
    \item \textbf{Plant}: Contains only plant samples.
    \item \textbf{Flower\_Pot}: Includes both the plant and the pot.
    \item \textbf{Vase}: Restricts objects to empty pots without plants.
    \item \textbf{Cup}: Consists of cups with handles.
    \item \textbf{Bowl}: Encompasses wide, low-height hemispherical shapes.
\end{itemize}

\begin{figure}[t]
    \centering
    \small
    \begin{subfigure}[b]{\columnwidth}
        \centering
        \includegraphics[width=0.3\linewidth]{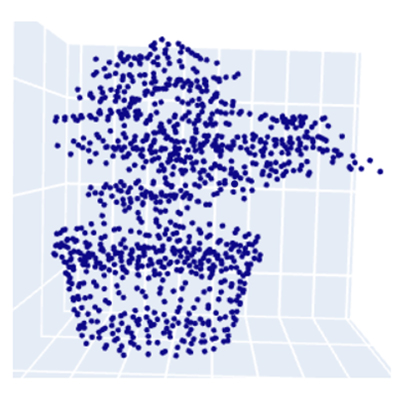}
        \includegraphics[width=0.3\linewidth]{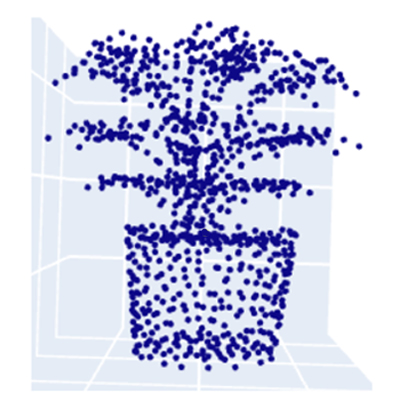}
        \includegraphics[width=0.3\linewidth]{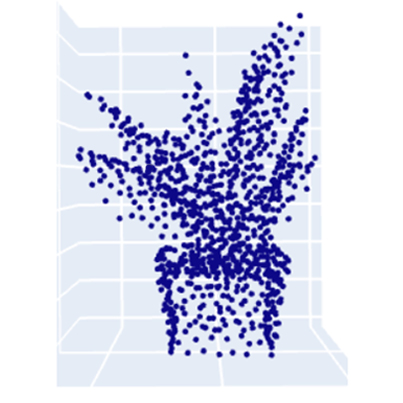}
        \caption{\textit{flower\_pot}}
        \label{subfig:row1}
    \end{subfigure}
    \vspace{1em}
    \begin{subfigure}[b]{\columnwidth}
        \centering
        \includegraphics[width=0.3\linewidth]{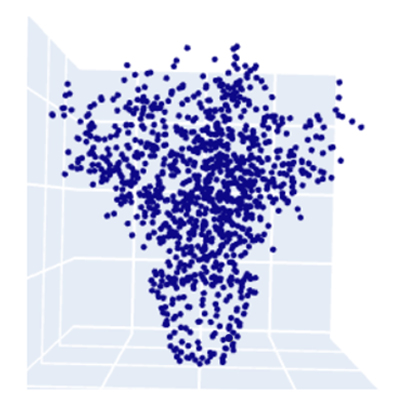}
        \includegraphics[width=0.3\linewidth]{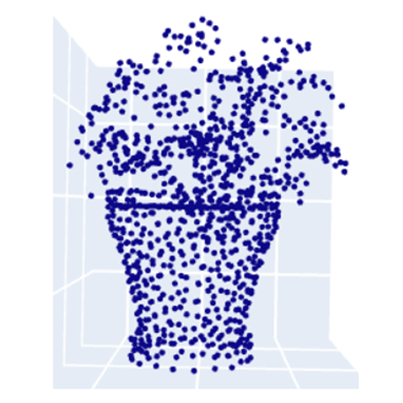}
        \includegraphics[width=0.3\linewidth]{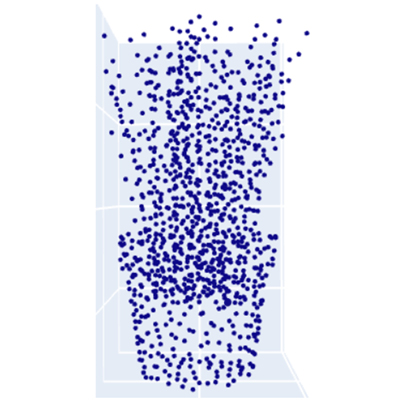}
        \caption{\textit{plant}}
        \label{subfig:row2}
    \end{subfigure}
    \vspace{-8mm}
    \caption{\small Examples of similar geometries in ModelNet classes.}
    \label{fig:similar1}
    \vspace{-5mm}
\end{figure}

\begin{figure}[t] 
    \centering
    \begin{subfigure}[b]{0.48\columnwidth}
        \centering
        \includegraphics[width=0.485\textwidth]{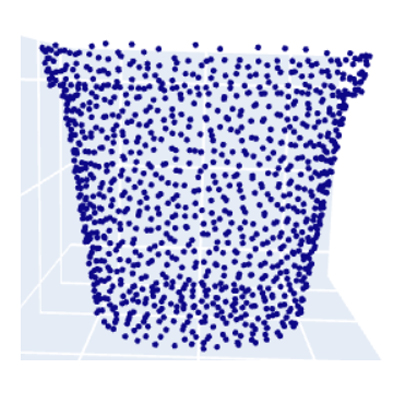}
        \includegraphics[width=0.485\textwidth]{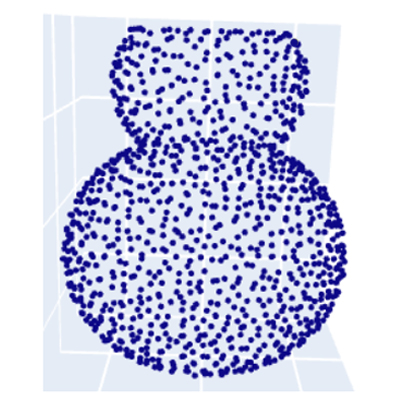}
        \caption{\textit{Flower\_pot}}
    \end{subfigure}
    \hfill
    \begin{subfigure}[b]{0.48\columnwidth}
        \centering
        \includegraphics[width=0.485\textwidth]{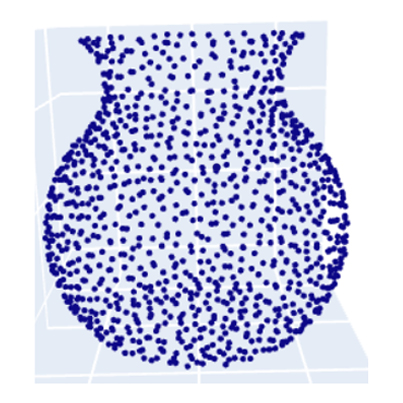}
        \includegraphics[width=0.485\textwidth]{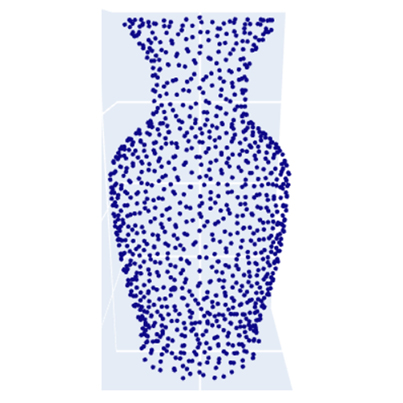}
        \caption{\textit{Vase}}
    \end{subfigure}
    \caption{Comparison of vase and flower pot with high surface similarity.}
    \label{fig:similar2}
    \vspace{-7mm}
\end{figure}

\autoref{fig:similar1} and \autoref{fig:similar2} demonstrate examples of these closely related classes, underscoring the importance of precise, non-overlapping definitions.

\subsection{Data Refinement Process}
\label{subsec:data_refine_process}

We used visual observations and confusion matrix analyzes to target classes with frequent misclassifications. For these classes, we meticulously moved or removed data samples based on the refined criteria. \autoref{table:data_refinement} provides an overview of these adjustments:

\begin{itemize}
    \item \textbf{Moved Instances:} Samples reassigned to a more appropriate class (e.g., reclassifying certain '\textit{plant}' instances as '\textit{flower\_pot}').
    \item \textbf{Removed Instances:} Ambiguous or low-quality samples eliminated entirely.
\end{itemize}

The 'Total' column in \autoref{table:data_refinement} shows the original sample count for each class, while the final row displays the updated counts in \textbf{ModelNet-R}. For example, in the class of '\textit{flower\_pot}', 72 samples were reclassified as '\textit{vase}', 5 as '\textit{bowl}' and 1 was removed. Meanwhile, 171 '\textit{plant}' samples were relabeled as '\textit{flower\_pot}', resulting in a final count of 262 samples in that class.

By rectifying mislabeled items, removing 2D data, adjusting class definitions, and addressing size mismatches, \textbf{ModelNet-R} offers a cleaner, more coherent dataset. These refinements are crucial for improving \textit{3D point cloud classification} and ensuring that future models trained on \textbf{ModelNet-R} produce more reliable and interpretable results.

\begin{table}[h]
    \centering
    \footnotesize
    \setlength{\tabcolsep}{3pt} 
    \renewcommand{\arraystretch}{1.1}
    \caption{Summary of modifications in ModelNet-R.}
    \label{table:data_refinement}
    \begin{tabular}{|c|c|c|c|c|c|c|c|}
        \hline
        Class & Flower & Plant & Vase & Cup & Bowl & Removed & Total \\
        & \_Pot &  & & & & & \\ \hline
        Flower\_Pot & 91 & 0 & 72 & 0 & 5 & 1 & 169 \\
        \hline
        Plant & 171 & 152 & 0 & 0 & 0 & 16 & 339 \\ \hline
        Vase & 0 & 0 & 571 & 0 & 2 & 2 & 575 \\ \hline
        Cup & 0 & 0 & 55 & 43 & 1 & 0 & 99 \\ \hline
        Bowl & 0 & 0 & 24 & 0 & 60 & 0 & 84 \\ \hline
        Total & 262 & 152 & 722 & 43 & 68 & 19 & 1266 \\ \hline
    \end{tabular}
    \vspace{-3mm}
\end{table}

\subsection{Point-SkipNet: A Lightweight Graph-Based Model for Point Cloud Classification}

\label{subsec:point_skipnet_overview}

This section presents \textbf{Point-SkipNet}, a lightweight graph-based architecture designed for efficient and accurate 3D point cloud classification.

\subsubsection{Background}
\label{subsubsec:point_skipnet_background}

A point cloud can be represented as a set of points in the 3D space. Each point has coordinates \((x, y, z)\in \mathbb{R}^3\). Formally, a point cloud \(P\) with \(N\) points can be written as \(P = \{p_i\}_{i=1}^{N} \subset \mathbb{R}^{3}\), where each point \(p_i\) is:

\begin{equation}
p_i = (x_i, y_i, z_i).
\label{equation:point_coordinate}
\end{equation}

Point cloud classification involves learning a function \(\text{cls}\) that maps \(P\) to a class label \(C\):

\begin{equation}
C = \operatorname{CLS}(P),
\label{equation:classification}
\end{equation}

During training, \(\text{CLS}\) is optimized to maximize classification accuracy based on extracted geometric and contextual features.
\begin{figure*}[t]
    \centering
        \includegraphics[width=1.5\columnwidth]{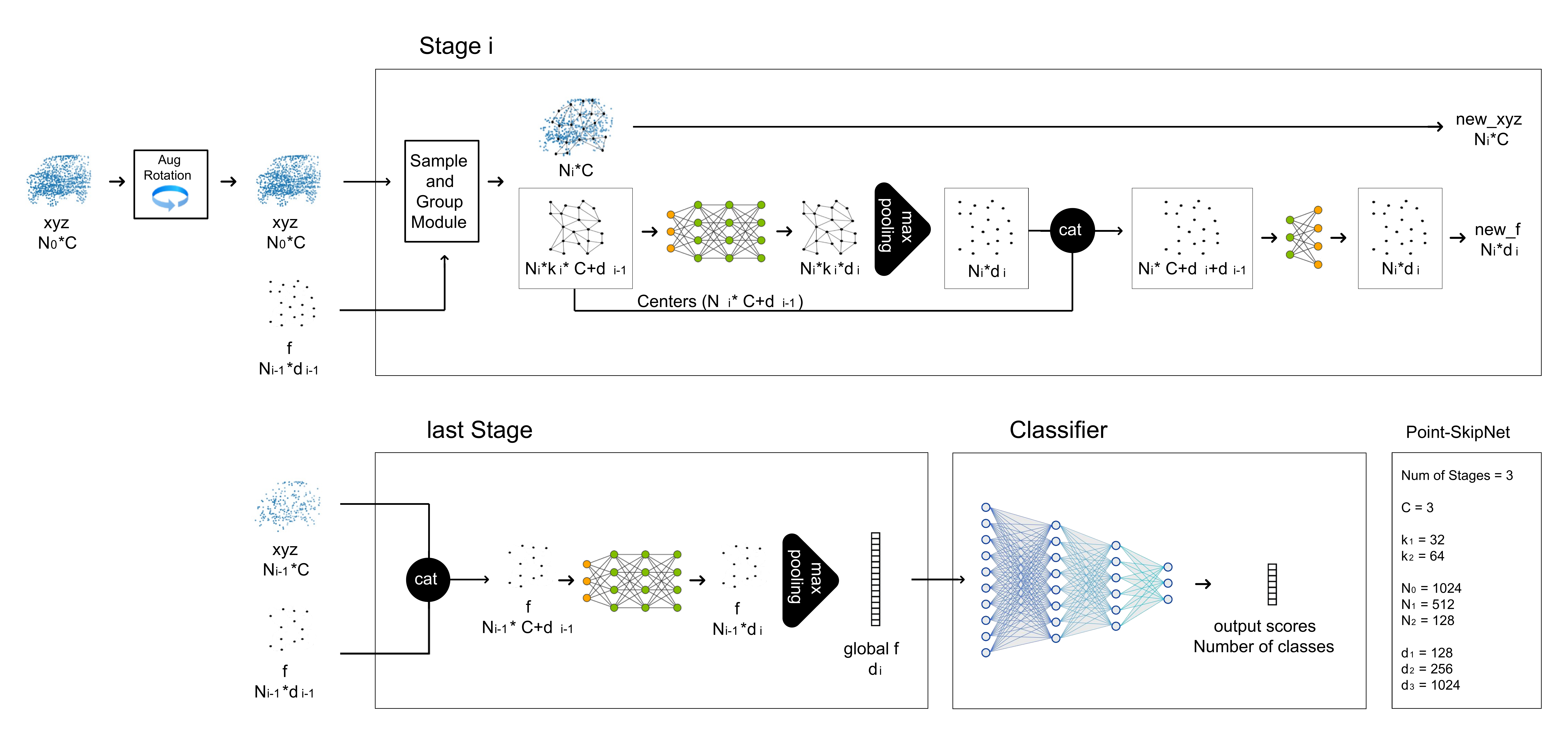}
        \caption{Point-SkipNet architecture.}
        \label{fig:Point-SkipNet}
\end{figure*}

\subsubsection{Point-SkipNet Architecture}
\label{subsubsec:point_skipnet_model}

\autoref{fig:Point-SkipNet} provides an overview of the \textbf{Point-SkipNet} architecture. The model begins by applying an augmentation function, denoted as \( \text{Aug}(\cdot) \), which may include rotations or other transformations:

\begin{equation}
    P_{\text{aug}} = \text{Aug}(P)
    \label{equation:augmentation}
\end{equation}

\noindent where \( P_{\text{aug}} \) is the augmented point cloud used in the subsequent stages. The classification layers then operate on learned global features to produce class scores. The following sections describe the primary components of \textbf{Point-SkipNet}.

\begin{figure*}[t]
    \centering
        \includegraphics[width=2\columnwidth]{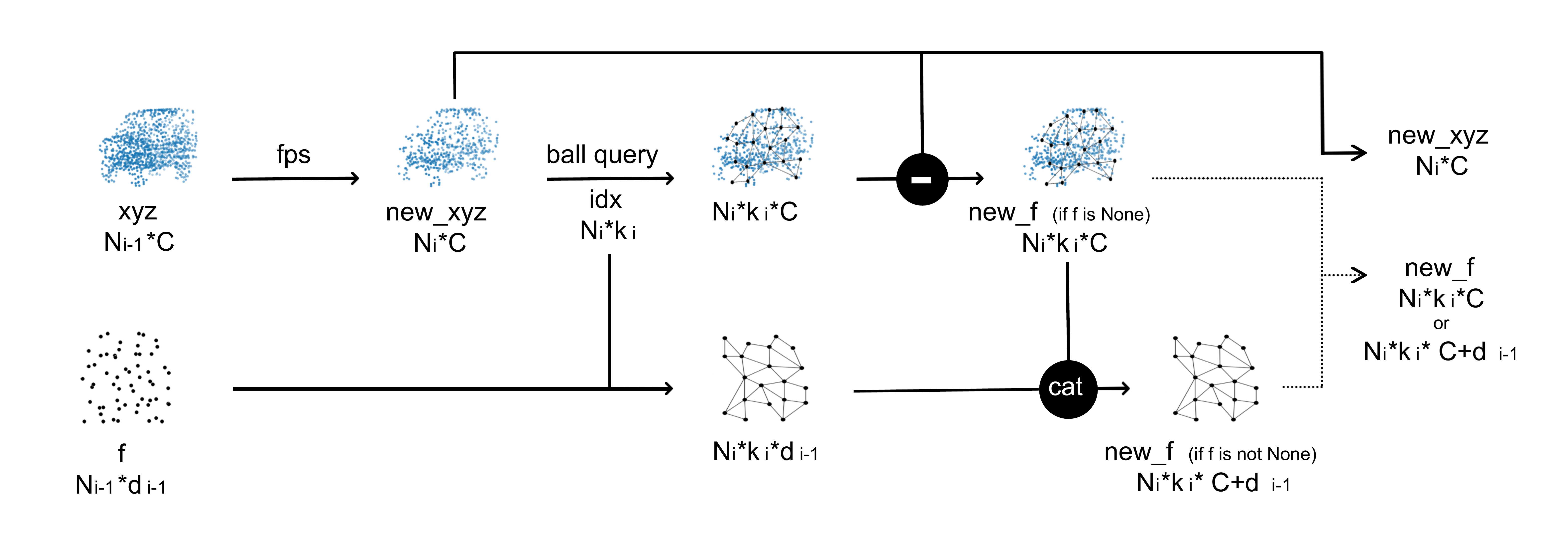}
        \caption{Sample and Group Module in Point-SkipNet.}
        \label{fig:Sample_and_group}
        \vspace{-5mm}
\end{figure*}

\textbf{Sample and Group Module.}
As illustrated in \autoref{fig:Sample_and_group}, the \textit{Sample and Group Module} begins by applying Farthest Point Sampling (FPS) to select a subset of points that uniformly covers the entire point cloud:

\begin{equation}
    \{ p_j \}_{j=1}^{N_j} = \text{FPS}(\{ p_i \}_{i=1}^{N_i}), \quad N_j < N_i
    \label{equation:sampling}
\end{equation}

After sampling, a \textit{ball query} is used to group each sampled point \( p_j \) with its \( k_j \) nearest neighbors within a predefined radius \( r \):

\begin{equation}
    G_j = \{ p_i \mid \| p_i - p_j \| \leq r \}, \quad \forall j \in \{1, \dots, N_j\}
    \label{equation:grouping}
\end{equation}

Here, \( G_j \) represents the set of neighboring points within the ball query radius \( r \), centered at each sampled point \( p_j \). This grouping ensures that local geometric structures are preserved for subsequent feature aggregation.

The full grouped representation of the sampled points is given by:

\begin{equation}
    G_P = \{ G_j \}_{j=1}^{N_j} = \text{ball\_query}(\{ p_i \}_{i=1}^{N_i}, \{ p_j \}_{j=1}^{N_j}, r)
    \label{equation:grouping_points}
\end{equation}

\noindent where \( G_P \in \mathbb{R}^{N_j \times k_j \times c} \), with \( k_j \) denoting the number of neighbors for each sampled point, and \( c \) representing the dimensionality of the point/feature channels.

If additional feature vectors \( f = \{ f_i \}_{i=1}^{N_i} \) are available, they are retrieved using the ball query indices:

\begin{equation}
    G_F = \text{gather}(f, G_P)
    \label{equation:features}
\end{equation}

Here, \( G_F \) represents the extracted features, and \( \text{gather}(\cdot) \) denotes an indexing operation that retrieves feature vectors for the neighbors defined by the ball query.

\textbf{Processing Stages.}
Each \textit{stage} in Point-SkipNet follows these steps:

\begin{enumerate}[label=(\roman*)]
    \item \textbf{Sample and Group:} Uses FPS and ball queries to collect local neighborhoods.
    \item \textbf{Feature Extraction:} Applies an MLP to transform the grouped features:
    \begin{equation}
        F_{\text{mlp}} = \text{MLP}(F_{\text{in}})
        \label{equation:feature_extraction_MLP}
    \end{equation}
    where \( F_{\text{in}} \in \mathbb{R}^{N_i \times k_i \times (c + d_{i-1})} \) is the input feature tensor, and \( F_{\text{mlp}} \in \mathbb{R}^{N_i \times k_i \times d_i} \) is the transformed feature set.
    
    \item \textbf{Symmetric Function:} A max-pooling operation reduces each neighborhood to a single feature vector:
    \begin{equation}
        F_{\text{pooled}} = \text{max\_pool}(F_{\text{mlp}}) \in \mathbb{R}^{N_i \times d_i}
    \label{equation:maxPooling}
    \end{equation}

    \item \textbf{Skip Connection:} The pooled features are concatenated with the center points to preserve original spatial information:
    \begin{equation}
        F_{\text{skip}} = \text{concat}(F_{\text{in}}, F_{\text{pooled}})
        \label{equation:concat}
    \end{equation}

    \item \textbf{Dimension Reduction:} A final MLP reduces the dimensionality:
    \begin{equation}
        F_{\text{out}} = \text{MLP}'(F_{\text{skip}})
        \label{equation:red_dim}
    \end{equation}
    where \( F_{\text{out}} \in \mathbb{R}^{N_i \times d_i} \) is the output feature set for the stage.
\end{enumerate}

\textbf{Final Stage and Global Feature.} 
In the final stage, the network aggregates all learned features to produce a \textbf{global descriptor} of the point cloud:

\begin{equation}
    F_{\text{global}} = \text{max\_pool}\!\Big(\text{MLP}''\big(\text{concat}(xyz, F)\big)\Big),
    \label{equation:final_stage}
\end{equation}

\noindent where \( F_{\text{global}} \in \mathbb{R}^{d_i} \) captures the overall characteristics of the point cloud.

\subsection{Classification Layer}
\label{subsubsec:point_skipnet_classifier}

The final classifier takes the \textbf{global feature descriptor} \( F_{\text{global}} \) as input and applies \textbf{fully connected (FC) layers} to produce class scores. If the dataset has \( K \) classes, the final output is:

\begin{equation}
    \text{scores} = \text{FC}(F_{\text{global}}) \in \mathbb{R}^K
    \label{equation:classifier}
\end{equation}

By integrating \textbf{efficient sampling, local feature aggregation, skip connections, and dimensionality reduction}, \textbf{Point-SkipNet} achieves high classification accuracy while maintaining computational efficiency. Its modular design allows for easy adaptation to various \textbf{3D vision tasks}, making it a powerful framework for \textbf{point cloud analysis}.

\section{Experiments}
\label{sec:experiments}
\begin{table*}[t]
\footnotesize
\centering
\caption{\small Performance Comparison of Various 3D Point Cloud Classification Models on ModelNet and ModelNet-R Datasets}
\label{table:combined_results}
\begin{tabular}{|l|c|c|c|c|c|c|c|}
\hline
\textbf{Model} & \multicolumn{2}{c|}{\textbf{Original Dataset}} & \multicolumn{2}{c|}{\textbf{ModelNet-R}} & \multicolumn{2}{c|}{\textbf{Performance Improvement}} & \textbf{Parameters (M)} \\ \hline
 & OA (\%) & mAcc (\%) & OA (\%) & mAcc (\%) & $\Delta$ OA (\%) & $\Delta$ mAcc (\%) & \\ \hline
PointNet~\cite{qi2017pointnet} & 89.20 & 86.00 & 91.39 & 88.79 & +2.19 & +2.79 & 3.47 \\ \hline
PointNet++ (SSG)~\cite{qi2017pointnet++} & - & - & 94.02 & 92.40 & +1.91 & +4.16 & 1.47 \\ \hline
PointNet++ (MSG)~\cite{qi2017pointnet++} & 90.70 & - & 94.06 & 91.80 & +3.36 & +1.80 & 1.74 \\ \hline
Point-NN~\cite{xiang2021walk} & 81.80 & - & 84.75 & 77.65 & +3.95 & +2.58 & 0.00 \\ \hline
DG-CNN~\cite{wang2019dynamic} & 92.90 & 90.20 & 94.03 & 92.64 & +1.13 & +2.44 & 1.80 \\ \hline
CurveNet~\cite{zhang2023parameter} & 93.80 & - & 94.12 & 92.65 & +0.32 & +2.70 & 2.04 \\ \hline
PointMLP~\cite{ma2022rethinking} & 94.10 & 91.10 & 95.33 & 94.30 & +1.23 & +3.20 & 12.60 \\ \hline
\rowcolor{lightgreen} \textbf{Point-SkipNet (Proposed)} & 92.29 & 89.84 & 94.33 & 92.93 & +2.04 & +3.09 & 1.47 \\ \hline
\end{tabular}
\vspace{-5mm}
\end{table*}

This section presents the experimental setup, datasets, parameter settings, and results of evaluating \textbf{Point-SkipNet}. We assess its performance on both ModelNet and the refined ModelNet-R, highlighting the impact of dataset refinement.


\subsection{Experimental Setup}

Experiments were conducted on an NVIDIA GeForce RTX 3080 GPU to efficiently train and evaluate models on 3D point cloud data. We used both ModelNet and its refined version, ModelNet-R, to analyze the benefits of dataset refinement.


\subsection{Datasets}

ModelNet is a benchmark dataset with 12,311 CAD models across 40 categories, widely used in 3D object classification. ModelNet-R is a refined version addressing label inconsistencies, 2D data artifacts, size mismatches, and class differentiation issues, leading to better training and evaluation reliability.

\subsection{Parameter Settings}
The models were trained with the following parameters: Batch Size = 32, Epochs = 200, Optimizer = Adam, Learning Rate = 0.001. These settings were chosen based on best practices in 3D classification.




\subsection{Evaluation Metrics}

We used two metrics: (i) Overall Accuracy (OA), which measures correct classifications as a percentage of total samples, and (ii) Mean Class Accuracy (mAcc), which calculates the average per-class accuracy to measure performance consistency.


\subsection{Results and Discussion}

We conducted a comparative analysis of \textbf{Point-SkipNet} against state-of-the-art models using both ModelNet and ModelNet-R. \autoref{table:combined_results} illustrates the performance of various models on the both datasets. Notably, \textbf{Point-SkipNet} achieves an \textit{Overall Accuracy (OA) of 92.29\%} and a \textit{Mean Class Accuracy (mAcc) of 89.84\%} on ModelNet. Its performance improves significantly on ModelNet-R, achieving \textit{94.33\% OA} and \textit{92.93\% mAcc}, reflecting an improvement of \textit{+2.04\% OA} and \textit{+3.09\% mAcc}. This result underscores both the effectiveness of the dataset refinement process and the efficiency of the \textbf{Point-SkipNet} architecture.

The following key observations were made based on the results of the experiments:
\begin{enumerate}[label=(\roman*)]
    \item \textbf{Impact of Dataset Refinement}: All models exhibit improved performance on ModelNet-R, demonstrating the importance of high-quality datasets in enhancing classification accuracy.
    \item \textbf{Efficiency of Point-SkipNet}: Despite its strong accuracy, \textbf{Point-SkipNet} maintains a relatively low parameter count (1.47M), making it suitable for deployment in resource-constrained environments.
    \item \textbf{Comparative Performance}: \textbf{Point-SkipNet} outperforms several existing models in terms of improvement rate, highlighting its robustness and efficiency.
\end{enumerate}

Additionally, we conducted ablation studies to analyze the impact of data augmentation and skip connection modes on the model’s performance. These results are presented in \autoref{table:augmentation} and \autoref{table:skip_connection}.

\subsection{Ablation Studies}

To further examine \textbf{Point-SkipNet's} design choices, we conducted ablation studies focusing on data augmentation techniques and skip connection modes.

\subsubsection{Data Augmentation}

We investigated different data augmentation strategies to enhance variability and improve model generalization. \autoref{table:augmentation} summarizes the results.

\begin{itemize}
    \item Among all augmentation techniques, \textbf{rotation augmentation} yielded the highest accuracy, indicating its importance in learning \textbf{rotational invariance}.
    \item Applying \textbf{all augmentations} simultaneously did not yield significant improvements, suggesting that certain augmentations may introduce redundancy.
\end{itemize}

\subsubsection{Skip Connection Modes}

We compared two skip connection strategies: concatenation and addition. \autoref{table:skip_connection} shows that concatenation consistently outperforms addition in both Overall Accuracy (OA) and Mean Class Accuracy (mAcc). This suggests that concatenating features retains richer information, leading to better classification performance.

\subsection{Summary and Implications}

Our experimental results demonstrate the effectiveness of both the refined ModelNet-R dataset and the proposed \textbf{Point-SkipNet model} in achieving superior 3D point cloud classification performance, as following:


\begin{enumerate}[label=(\roman*)]
    \item \textbf{Dataset Quality Matters}: Refining data by correcting labels, removing low-quality samples, and improving class differentiation significantly enhances model accuracy.
    \item \textbf{Model Efficiency}: \textbf{Point-SkipNet} delivers competitive accuracy while maintaining a lightweight architecture, making it suitable for deployment in resource-constrained environments such as \textbf{mobile devices and embedded systems}.
    \item \textbf{Architectural Design Choices}: The use of concatenation-based skip connections and strategic data augmentation play crucial roles in improving classification performance.
\end{enumerate}

These findings highlight the importance of high-quality datasets and efficient model architectures in advancing 3D point cloud classification. Future work will explore further optimizations to \textbf{Point-SkipNet} and extend dataset refinement efforts to other benchmarks.

\begin{table}[t]

\centering
\footnotesize
\caption{Impact of Data Augmentation Techniques}
\label{table:augmentation}
\begin{tabular}{|l|c|c|}
\hline
\textbf{Augmentation Mode} & \textbf{OA (\%)} & \textbf{mAcc (\%)} \\ \hline
Main & 93.72 & 92.56 \\ \hline
All Augmentations & 93.49 & 92.25 \\ \hline
Anisotropic Scaling & 93.76 & 92.59 \\ \hline
Jitter & 93.56 & 92.19 \\ \hline
Rotation & 93.93 & 92.55 \\ \hline
Translation & 93.75 & 92.49 \\ \hline
\end{tabular}
\vspace{-2mm}
\end{table}

\begin{table}[t]

\centering
\footnotesize
\caption{Effect of Skip Connection Modes}
\label{table:skip_connection}
\begin{tabular}{|l|c|c|}
\hline
\textbf{Skip Connection Mode} & \textbf{OA (\%)} & \textbf{mAcc (\%)} \\ \hline
Concatenation & 93.79 & 92.49 \\ \hline
Addition & 93.64 & 92.09 \\ \hline
\end{tabular}
\vspace{-4mm}
\end{table}

\section{Limitations and Future Work}
\label{sec:limit_future}

This study advances 3D point cloud classification but has limitations that suggest future research directions.

Firstly, refinement was applied to only 5 of 40 ModelNet40~\cite{Wu_2015_CVPR} classes. Expanding this to all classes will ensure dataset consistency.

Secondly, data normalization removed size-related information. Future research should explore techniques that retain this, such as incorporating size ratios.

Thirdly, \textbf{Point-SkipNet} was only tested on ModelNet and ModelNet-R. Further validation on diverse datasets is needed, along with reevaluating models using ModelNet-R for fair comparisons.

Future work includes refining all ModelNet40 classes, preserving size-related data, evaluating \textbf{Point-SkipNet} across datasets, and reassessing models on ModelNet-R. Addressing these will enhance dataset quality and model performance in 3D point cloud analysis.

\section{Conclusion}
\label{sec:conclusion}

In this work, we addressed key challenges in 3D point cloud classification by introducing ModelNet-R, a refined benchmark dataset, and \textbf{Point-SkipNet}, a lightweight graph-based neural network. ModelNet-R mitigates labeling inconsistencies, removes low-quality data, and improves class differentiation, leading to a more reliable evaluation framework. Our extensive experiments demonstrate that Point-SkipNet achieves state-of-the-art accuracy while maintaining a significantly lower computational footprint, making it suitable for real-time and resource-constrained applications.

Furthermore, our analysis highlights the critical interplay between dataset quality and model efficiency in advancing 3D perception tasks. The proposed data refinement process improves classification accuracy across multiple models, reinforcing the necessity of high-quality benchmarks in deep learning research.

Future work will focus on extending the dataset refinement process to all ModelNet40 classes and incorporating real-world noisy datasets to enhance generalization. Additionally, we plan to explore advanced normalization techniques that retain size-related information and validate \textbf{Point-SkipNet} on diverse 3D benchmarks such as ScanObjectNN and ShapeNet. By emphasizing both efficient model design and dataset integrity, this research contributes to the broader goal of developing accurate and computationally efficient 3D classification models, paving the way for improved applications in robotics, autonomous driving, and augmented reality.




\bibliographystyle{unsrt}


\end{document}